\documentclass[runningheads]{llncs}
\usepackage{graphicx}

\usepackage{tikz}
\usepackage{comment}
\usepackage{amsmath,amssymb} 
\usepackage{color}

\usepackage{graphicx}
\usepackage{multirow}
\usepackage{multicol}

\usepackage{makecell}
\usepackage{subfigure}
\usepackage{longtable}

\usepackage{epsfig}
\usepackage{amsmath}
\usepackage{amssymb}
\usepackage{bm}
\usepackage{mathtools}
\usepackage{array}
\usepackage{colortbl}
\usepackage{booktabs}
\usepackage{bbm}
\usepackage{xcolor}
\usepackage{xspace}
\usepackage{float}
\usepackage{pifont}
\usepackage{wrapfig}
\usepackage{threeparttable}
\usepackage{footnote}
\usepackage{overpic}
\usepackage{url}

\usepackage{pdfcomment}

\usepackage{listings}

\definecolor{linkcolor}{RGB}{255,0,0}
\definecolor{urlcolor}{RGB}{255,105,180}
\definecolor{citecolor}{RGB}{66,168,235}
\definecolor{codegreen}{rgb}{0,0.5,0}
\definecolor{codeblue}{rgb}{0.25,0.5,0.5}
\definecolor{codegray}{rgb}{0.6,0.6,0.6}

\usepackage{hyperref}
\hypersetup{colorlinks=true,linkcolor=linkcolor,urlcolor=urlcolor,citecolor=citecolor}

\definecolor{mygray}{gray}{.93}
\definecolor{mygray1}{gray}{.99}

\newcommand{\tabincell}[2]{\begin{tabular}{@{}#1@{}}#2\end{tabular}}

\makeatletter
\DeclareRobustCommand\onedot{\futurelet\@let@token\@onedot}
\def\@onedot{\ifx\@let@token.\else.\null\fi\xspace}

\def\eg{\emph{e.g}\onedot} 
\def\ie{\emph{i.e}\onedot} 
 
\def\etc{\emph{etc}\onedot} 
 
\def\etal{\emph{et al}\onedot}
\makeatother

\begin{document}

\pagestyle{headings}
\mainmatter
\def\ECCVSubNumber{1174}  

\makeatletter
\def\@fnsymbol#1{\ensuremath{\ifcase#1\or *\or \dagger\or \ddagger\or
   \mathsection\or \mathparagraph\or \|\or **\or \dagger\dagger
   \or \ddagger\ddagger \else\@ctrerr\fi}}
    \makeatother
    
\title{Towards Data-Efficient Detection Transformers} 

\titlerunning{Towards Data-Efficient Detection Transformers}

\author{
Wen Wang\inst{1}\thanks{This work was done during Wen Wang’s internship at JD Explore Academy.},
Jing Zhang\inst{2}\thanks{Co-first author.},
Yang Cao\inst{1,3}\thanks{Corresponding author.},
Yongliang Shen\inst{4},
Dacheng Tao\inst{5,2}
}

\authorrunning{W. Wang et al.}
%
\institute{
$^{1}$University of Science and Technology of China \quad
$^{2}$The University of Sydney \quad
$^{3}$Institute of Artificial Intelligence, Hefei Comprehensive National Science Center \quad
$^{4}$Zhejiang University \quad
$^{5}$JD Explore Academy, China \\
\email{
wangen@mail.ustc.edu.cn \quad
jing.zhang1@sydney.edu.au \\
forrest@ustc.edu.cn \quad
syl@zju.edu.cn \quad
dacheng.tao@gmail.com
}
}
\maketitle

\begin{abstract}
Detection transformers have achieved competitive performance on the sample-rich COCO dataset. However, we show most of them suffer from significant performance drops on small-size datasets, like Cityscapes. In other words, the detection transformers are generally data-hungry. To tackle this problem, we empirically analyze the factors that affect data efficiency, through a step-by-step transition from a data-efficient RCNN variant to the representative DETR. The empirical results suggest that sparse feature sampling from local image areas holds the key. Based on this observation, we alleviate the data-hungry issue of existing detection transformers by simply alternating how key and value sequences are constructed in the cross-attention layer, with minimum modifications to the original models. Besides, we introduce a simple yet effective label augmentation method to provide richer supervision and improve data efficiency. Experiments show that our method can be readily applied to different detection transformers and improve their performance on both small-size and sample-rich datasets. Code will be made publicly available at \url{https://github.com/encounter1997/DE-DETRs}.

\end{abstract}

\section{Introduction}
\label{sec:intro}

Object detection is a long-standing topic in computer vision. 
Recently, a new family of object detectors, named detection transformers, has drawn increasing attention due to their simplicity and promising performance. The pioneer work of this class of methods is DETR~\cite{detr}, which views object detection as a direct set prediction problem and applies a transformer to translate the object queries to the target objects. It achieves better performance than the seminal Faster RCNN~\cite{faster-rcnn} on the commonly used COCO dataset~\cite{coco}, but its convergence is significantly slower than that of CNN-based detectors. For this reason, most of the subsequent works have been devoted to improving the convergence of DETR, through efficient attention mechanism~\cite{zhu2020deformable}, conditional spatial query~\cite{meng2021conditional}, regression-aware co-attention~\cite{smca}, \etc. These methods are able to achieve better performance than Faster RCNN with comparable training costs on the COCO dataset, demonstrating the superiority of detection transformers.

\begin{figure*}[t]
  \centerline{\includegraphics[width=10cm]{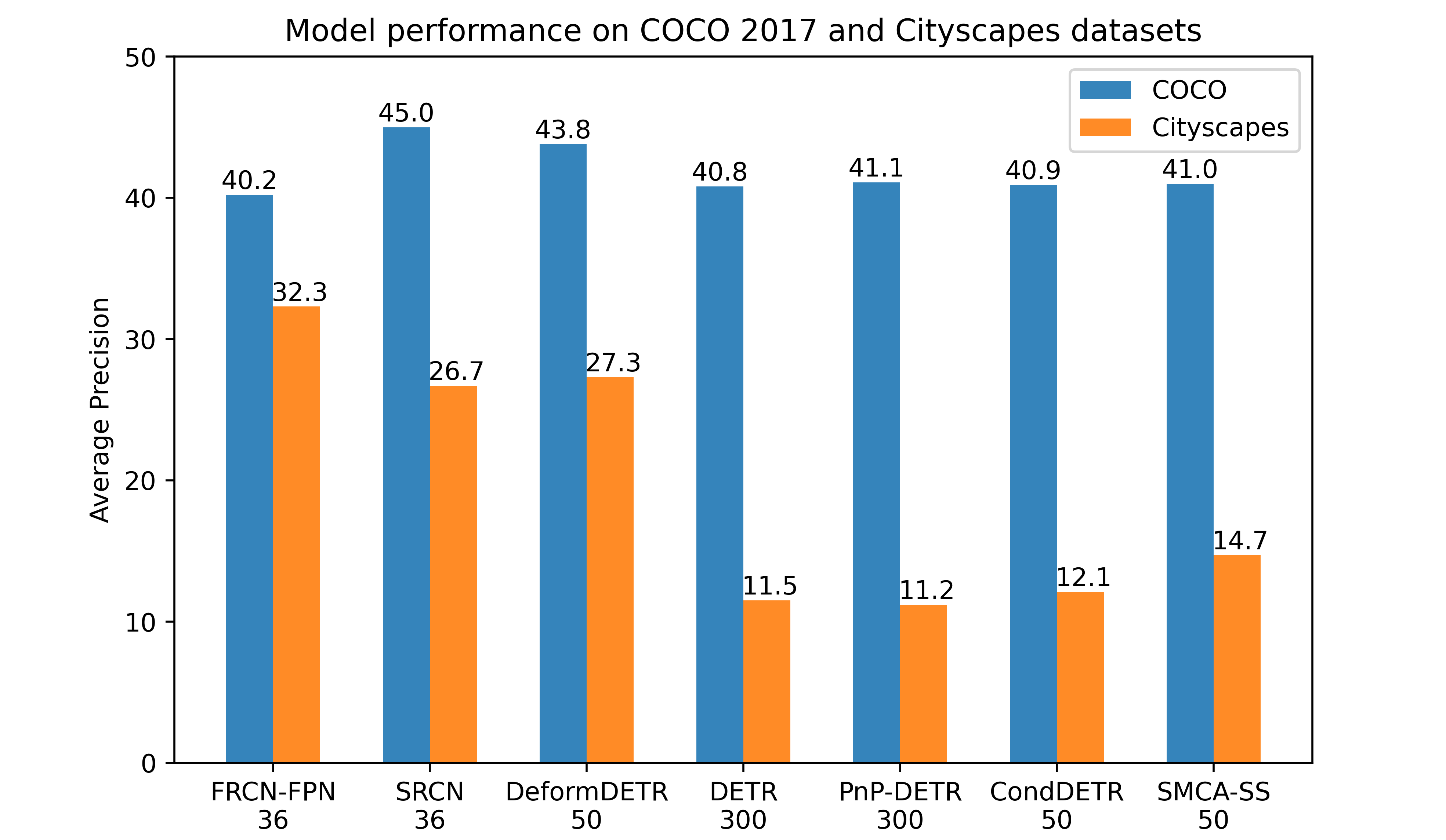}}
  \caption{Performance of different object detectors on COCO 2017 with 118K training data and Cityscapes with 3K training data. The respective training epochs are shown below the name of each method. While the RCNN family show consistently high average precision, the detection transformer family degrades significantly on the small-size dataset. 
  FRCN-FPN, SRCN, and SMCA-SS represent Faster-RCNN-FPN, Sparse RCNN, and single-scale SMCA, respectively. 
  }
  \label{fig:bar}
\end{figure*}

Current works seem to suggest that detection transformers are superior to the CNN-based object detector, like Faster RCNN, in both simplicity and model performance. However, we find that detection transformers show superior performance only on datasets with rich training data like COCO 2017 (118K training images), while the performance of most detection transformers drops significantly when the amount of training data is small. For example, on the commonly used autonomous driving dataset Cityscapes~\cite{cityscapes} (3K training images), the average precisions (AP) of most of the detection transformers are less than half of Faster RCNN AP performance, as shown in Fig.~\ref{fig:bar}.
Moreover, although the performance gaps between different detection transformers on the COCO dataset are less than 3 AP, a significant difference of more than 15 AP exists on the small-size Cityscapes dataset.

These findings suggest that detection transformers are generally more data-hungry than CNN-based object detectors. However, the acquisition of labeled data is time-consuming and labor-intensive, especially for the object detection task, which requires both categorization and localization of multiple objects in a single image. What’s more, the large amount of training data means more training iterations to traverse the dataset, and thus more computational resources are consumed to train the detection transformers, increasing the carbon footprint. In a word, it takes a lot of human labor and computational resources to meet the training requirements of existing detection transformers.

To address these issues, we first empirically analyze the key factors affecting the data efficiency of detection transformers through a step-by-step transformation from the data-efficient Sparse RCNN to the representative DETR. Our investigation and analysis show that sparse feature sampling from local area holds the key: on the one hand, it alleviates the difficulty of learning to focus on specific objects, and on the other hand, it avoids the quadratic complexity of modeling image features and makes it possible to utilize multi-scale features, which has been proved critical for the object detection task.

Based on these observations, we improve the data efficiency of existing detection transformers by simply alternating how the key and value are constructed in the transformer decoder. Specifically, we perform sparse sampling features on key and value features sent to the cross-attention layer under the guidance of the bounding boxes predicted by the previous decoder layer, with minimum modifications to the original model, and without any specialized module. In addition, we mitigate the data-hungry problem by providing richer supervisory signals to detection transformers. To this end, we propose a label augmentation method to repeat the labels of foreground objects during label assignment, which is both effective and easy to implement. Our method can be applied to different detection transformers to improve their data efficiency. Interestingly, it also brings performance gain on the COCO dataset with a sufficient amount of data.

To summarize, our contributions are listed as follows.
\begin{itemize}
    \item We identify the data-efficiency problem of detection transformers. Though they achieve excellent performance on the COCO dataset, they generally suffer from significant performance degradation on small-size datasets.
    \item We empirically analyze the key factor that affects detection transformers’ data efficiency through a step-by-step model transformation from Sparse RCNN to DETR, and find that sparse feature sampling from local areas holds the key to data efficiency.
    \item With minimum modifications, we significantly improve the data efficiency of existing detection transformers by simply alternating how key and value sequences are constructed in the cross-attention layer.
    \item We propose a simple yet effective label augmentation strategy to provide richer supervision and improve the data efficiency. It can be combined with different methods to achieve performance gains on different datasets.
\end{itemize}

\section{Related Work}
\label{sec:related_work}
\subsection{Object Detection}
Object detection~\cite{dpm,rcnn,faster-rcnn,yolo,ssd,lin2017focal,tian2019fcos} is essential to many real-world applications, like autonomous driving, defect detection, and remote sensing. 
Representative deep-learning-based object detection methods can be roughly categorized into two-stage detectors like Faster RCNN~\cite{faster-rcnn} and one-stage object detectors like YOLO~\cite{yolo} and RetinaNet~\cite{lin2017focal}. 
While effective, these methods generally rely on many heuristics like anchor generation and rule-based label assignments. 

Recently, DETR~\cite{detr} provides a simple and clean pipeline for object detection. It formulates object detection as a set prediction task, and applies a transformer~\cite{vaswani2017attention} to translate sparse object candidates~\cite{sparse-rcnn} to the target objects. The success of DETR has sparked the recent surge of detection transformers~\cite{zhu2020deformable,up-detr,smca,meng2021conditional,wb-detr,pnp-detr,yolos,wang2022fpdetr,chen2022recurrent,yuan2021polyphonicformer} and most of the following-up works focus on alleviating the slow convergence problem of DETR. 
For example, DeformDETR~\cite{zhu2020deformable} propose the deformable attention mechanism for learnable sparse feature sampling and aggregates multi-scale features to accelerate model convergence and improve model performance. 
CondDETR~\cite{meng2021conditional} proposes to learn a conditional spatial query from the decoder embedding, which helps the model quickly learn to localize the four extremities for detection.
    
These works achieve better performance than Faster RCNN on the COCO dataset~\cite{coco} with comparable training costs. It seems that detection transformers have surpassed the seminal Faster RCNN in both simplicity and superior performance. But we show that detection transformers are generally more data-hungry and perform much worse than Faster RCNN on small-size datasets.

\subsection{Label Assignment}
Label assignment~\cite{GuidedAnchor,MetaAnchor,FreeAnchor,autoassign,ota,shen2022acl} is a crucial component in object detection. It matches the ground truth of an object with a specific prediction from the model, and thereby provides the supervision signal for training. Prior to DETR, most object detectors~\cite{faster-rcnn,yolo,lin2017focal} adopt the one-to-many matching strategy, which assigns each ground truth to multiple predictions based on local spatial relationships. By contrast, DETR makes one-to-one matching between ground truths and predictions by minimizing a global matching loss. This label assignment approach has been followed by various subsequent variants of the detection transformer~\cite{zhu2020deformable,meng2021conditional,up-detr,yolos,wang2022fpdetr}. Despite the merits of avoiding the duplicates removal process, only a small number of object candidates are supervised by the object labels in each iteration. As a result, the model has to obtain enough supervised signals from a larger amount of data or more training epochs.

\subsection{Data-Efficiency of Vision Transformers}
Vision Transformers~\cite{dosovitskiy2020image,t2t,tnt,swin,pvt,xu2021vitae,zhang2022vitaev2,fang2021msg,chu2021twins} (ViTs) are emerging as an alternative to CNN for feature extractors and visual recognition. Despite the superior performance, they are generally more data-hungry than their CNN counterparts. 
To tackle this problem, DeiT~\cite{deit} improves its data efficiency by knowledge distillation from pre-trained CNNs, coupled with a better training recipe. 
Liu \etal propose a dense relative localization loss to improve ViTs’ data efficiency~\cite{liu2021efficient}.
Unlike the prior works~\cite{deit,liu2021efficient,cao2022training} that focus on the data efficiency issue of transformer backbones on image classification tasks, we tackle the data efficiency issue of detection transformers on the object detection task.

\section{Difference Analysis of RCNNs and DETRs}
\label{sec:model_transform}

As can be seen in Fig.~\ref{fig:bar}, detection transformers are generally more data-hungry than RCNNs. To find out the key factors to data efficiency, we transform a data-efficient RCNN step-by-step into a data-hungry detection transformer to ablate the effects of different designs. Similar research approach has also been adopted by ATSS~\cite{atss} and Visformer~\cite{chen2021visformer}, but for different research purposes.

\subsection{Detector Selection}
\begin{table*}[t]
\begin{center}
\setlength{\tabcolsep}{0.12cm}
\resizebox{\textwidth}{!}{
\begin{tabular}{l|c|c|c|c|c|c}
\Xhline{2\arrayrulewidth}
\hline
\rowcolor{mygray}
{Model} & {Added} & {Removed} & {50E AP} & {300E AP} & \tabincell{c}{Params} & \tabincell{c}{FLOPs} \\
\hline
SRCN & \multicolumn{2}{c|}{--} & 29.4 & 35.9 & 106M & 631G \\
\hline
Net1 & DETR Recipe & SRCN Recipe & 30.6 (\textcolor{blue}{+1.2}) & 34.4 (\textcolor{red}{-1.5}) & 106M & 294G \\
Net2 & -- & FPN & 23.3 (\textcolor{red}{-7.3}) & 26.6 (\textcolor{red}{-7.8}) & 103M & 244G \\
Net3 & transformer encoder & -- & 21.0 (\textcolor{red}{-2.3}) & 27.5 (\textcolor{blue}{+0.9}) & 111M & 253G \\ 
Net4 & cross-attn in decoder & dynamic conv & 18.1 (\textcolor{red}{-2.9}) & 25.4 (\textcolor{red}{-2.1}) & 42M & 86G \\ 
Net5 & dropout in decoder & -- & 16.7 (\textcolor{red}{-1.4}) & 26.1 (\textcolor{blue}{+0.7}) & 42M & 86G \\ 
Net6 & -- & bbox refinement & 15.0 (\textcolor{red}{-1.7}) & 22.7 (\textcolor{red}{-3.4}) & 41M & 86G \\
Net7 & -- & RoIAlign & 6.6 (\textcolor{red}{-8.4}) & 17.7 (\textcolor{red}{-5.0}) & 41M & 86G \\
\hline
DETR & -- & initial proposals & 1.6 (\textcolor{red}{-5.0}) & 11.5 (\textcolor{red}{-6.2}) & 41M & 86G \\
\Xhline{2\arrayrulewidth}
\hline
\end{tabular}
}
\end{center}
\caption{Model transformation from Sparse RCNN (SRCN for short) to DETR, experimented on Ciytscapes~\cite{cityscapes}. ``50E AP" and ``300E AP" indicate average precision after training for 50 and 300 epochs respectively. 
The change in AP is shown in the brackets, where \textcolor{red}{red} indicates drops and \textcolor{blue}{blue} indicates gains on AP.
}
\label{tab:transition}
\end{table*}
To obtain insightful results from the model transformation, we need to choose the appropriate detectors to conduct the experiments. To this end, we choose Sparse RCNN and DETR for the following reasons. Firstly, they are representative detectors from the RCNN and detection transformer families, respectively. The observations and conclusions drawn from the transformation between them shall also be helpful to other detectors.
Secondly, there is large difference between the two detectors in data efficiency, as shown in Fig.~\ref{fig:bar}. Thirdly, they share many similarities in label assignment, loss design, and optimization, which helps us eliminate the less significant factors while focus more on the core differences.

\subsection{Transformation from Sparse RCNN to DETR}
\label{subsec:transition}
During the model transformation, we consider two training schedules that are frequently used in detection transformers. The first is training for 50 epochs and learning rate decays after 40 epochs, denoted as 50E. And the second is training for 300 epochs and learning rate decays after 200 epochs. The transformation process is summarized in Table~\ref{tab:transition}. 

\noindent\textbf{Alternating training recipe.} 
Though Sparse RCNN and DETR share many similarities, there are still slight differences in their training Recipes, including the classification loss, the number of object queries, learning rate, and gradient clip. We first eliminate these differences by replacing the Sparse RCNN training recipe with the DETR training recipe. 
Eliminating the differences in training recipes helps us focus more on the key factors that affect the data-efficiency.

\noindent\textbf{Removing FPN.}
Multi-scale feature fusion has been proved effective for object detection~\cite{fpn}. 
The attention mechanism has a quadratic complexity with respect to the image scale, making the modeling of multi-scale features in DETR non-trivial. Thus DETR only takes 32$\times$ down-sampled single-scale feature for prediction. In this stage, we remove the FPN neck and send only the 32$\times$ down-sampled feature to the detection head, which is consistent with DETR. As expected, without multi-scale modeling, the model performance degrades significantly by 7.3 AP under the 50E schedule, as shown in Table~\ref{tab:transition}.

\noindent\textbf{Introducing transformer encoder.}
In DETR, the transformer encoder can be regarded as the neck in the detector, which is used to enhance the features extracted by the backbone. After removing the FPN neck, we add the transformer encoder neck to the model. 
It can be seen that the AP result decreases at 50E schedule while improves at 300E schedule. We conjecture that similar to ViT~\cite{dosovitskiy2020image}, the attention mechanism in the encoder requires longer training epochs to converge and manifest its advantages, due to the lack of inductive biases.

\noindent\textbf{Replacing dynamic convolutions with cross-attention.}
A very interesting design in Sparse RCNN is the dynamic convolution~\cite{jia2016dynamic,tian2020conditional} in the decoder, which acts very similar to the role of cross-attention in DETR. Specifically, they both adaptively aggregate the context from the image features to the object candidates based on their similarity. In this step, we replace the dynamic convolution with the cross-attention layer with learnable query positional embedding, and the corresponding results are shown in Table~\ref{tab:transition}. 
Counter-intuitively, a larger number of learnable parameters does not necessarily make the model more data-hungry. In fact, the dynamic convolutions with about 70M parameters can exhibit better data efficiency than the parameter-efficient cross-attention layer.

\noindent\textbf{Aligning dropout settings in the decoder.}
A slight difference between Sparse RCNN and DETR is the use of dropout layers in self-attention and FFN layers in the decoder. In this stage, we eliminate the interference of these factors. 

\noindent\textbf{Removing cascaded bounding box refinement.}
Sparse RCNN follows the cascaded bounding box regression in Cascade RCNN~\cite{cascade-rcnn}, where each decoder layer iteratively refines the bounding box predictions made by the previous layer. We remove it in this stage and as expected, the model performance degrades to some extent.

\noindent\textbf{Removing RoIAlign.}
Sparse RCNN, like other detectors in the RCNNs family, samples features from local regions of interest, and then makes predictions based on the sampled sparse features~\cite{sparse-rcnn}. By contrast, each content query in DETR aggregates object-specific information directly from the global features map. In this step, we remove the RoIAlign~\cite{mask-rcnn} operation in Sparse RCNN, with the box target transformation~\cite{rcnn}. 
It can be seen that significant degradation of the model performance occurs, especially under the 50E schedule, the model performance decreases by 8.4 AP. We conjecture that learning to focus on local object regions from the entire feature map is non-trivial. The model requires more data and training epochs to capture the locality properties. 

\noindent\textbf{Removing initial proposals.}
Finally, DETR directly predicts the target bounding boxes, while RCNNs make predictions relative to some initial guesses. In this step, we eliminate this difference by removing the initial proposal. Unexpectedly, this results in a significant decrease in model performance. We suspect that the initial proposal works as a spatial prior that helps the model to focus on object regions, thus reducing the need to learn locality from large training data.

\subsection{Summary}
By far, we have completed the model transformation from Sparse RCNN to DETR. From Table~\ref{tab:transition} and our analysis in Section~\ref{subsec:transition}, 
it can be seen that three factors result in more than 5 AP performance changes, and are key to data-efficient:
(a) sparse feature sampling from local regions, \eg, using RoIAlign; (b) multi-scale features which depend on sparse feature sampling to be computationally feasible; (c) prediction relative to initial spatial priors. Among them, (a) and (c) help the model to focus on local object regions and alleviate the requirement of learning locality from a large amount of data, while (b) facilitates a more comprehensive utilization and enhancement of the image features, though it also relies on sparse features.

It is worth mentioning that DeformDETR~\cite{zhu2020deformable} is a special case in the detection transformer family, which shows comparable data efficiency to Sparse RCNN. Our conclusions drawn from the Sparse RCNN to DETR model transformation can also explain DeformDETR’s data efficiency. Specifically, multi-scale deformable attention samples sparse features from local regions of the image and utilizes multi-scale features. The prediction of the model is relative to the initial reference points. Thus, all three key factors are satisfied in DeformDETR, though it was not intended to be data-efficient on small-size datasets.

\section{Method}
\label{sec:method}
In this section, we aim to improve the data efficiency of existing detection transformers, while making minimum modifications to their original designs. Firstly, we provide a brief revisiting of existing detection transformers. Subsequently, based on experiments and analysis in the previous section, we make minor modifications to the existing data-hungry detection transformer models, like DETR~\cite{detr} and CondDETR~\cite{meng2021conditional}, to significantly improve their data efficiency. Finally, we propose a simple yet effective label augmentation method to provide richer supervised signals to detection transformers to further improve their data efficiency.

\subsection{A Revisit of Detection Transformers}
\textbf{Model Structure.}
Detection transformers generally consist of a backbone, a transformer encoder, a transformer decoder, and the prediction heads. The backbone first extracts multi-scale features from the input image, denoted as $\left\{f^{l}\right\}_{l=1}^{L}$, where $f^{l} \in \mathbb{R}^{H^{l} \times W^{l} \times C^{l}}$. Subsequently, the last feature level with the lowest resolution is flattened and embedded to obtain $z^{L} \in \mathbb{R}^{S^{L} \times D}$ where $S^L=H^L \times W^L$ is sequence length and $D$ is the feature dimension. Correspondingly, the positional embedding is denoted as $p^L \in \mathbb{R}^{S^L \times D}$. Afterward, The single-scale sequence feature is encoded by the transformer encoder to obtain $z_{e}^{L} \in \mathbb{R}^{S^L \times D}$.

The decoder consists of a stack of $L_{d}$ decoder layers, and the query content embedding is initialized as $\mathbf{q}_{0} \in \mathbb{R}^{N \times D}$, where $N$ is the number of queries. Each decoder layer $\operatorname{DecoderLayer}_{\ell}$ takes the previous decoder layer’s output $\mathbf{q}_{\ell-1}$, the query positional embedding $p_{q}$, the image sequence feature $\mathbf{z}_{\ell}$ and its position embedding $p_{\ell}$ as inputs, and outputs the decoded sequence features.
\begin{equation}
\label{eq:decoder}
    \mathbf{q}_{\ell}=\operatorname{DecoderLayer}_{\ell}\left(\mathbf{q}_{\ell-1}, p_{q}, \mathbf{z}_{\ell}, p_{\ell}\right), 
    \quad \ell=1 \ldots L_{\mathrm{d}}.
\end{equation}
In most detection transformers, like DETR and CondDETR, single-scale image feature is utilized for decoder, and thus $\mathbf{z}_{\ell}=z_{e}^{L}$ and $p_{\ell}=p^{L}$, where $\ell=1 \ldots L_{\mathrm{d}}$.

\noindent\textbf{Label Assignment.}
Detection transformers view the object detection task as a set prediction problem and perform deep supervision~\cite{lee2015deeply} on predictions made by each decoder layer. Specifically, the labels set can be denoted as $y=\left\{y_{1}, \ldots, y_{M}, \varnothing, \ldots, \varnothing\right\}$, where $M$ denotes the number of foreground objects in the image and the $\varnothing$ (no object) pads the label set to a length of $N$. Correspondingly, the output of each decoder layer can be written as $\hat{y}=\left\{\hat{y}_{i}\right\}_{i=1}^{N}$. During label assignment, detection transformers search for a permutation $\tau \in T_{N}$ with the minimum matching cost:
\begin{equation}
    \hat{\tau}=\underset{\tau \in T_{N}}{\arg \min} \sum_{i}^{N} \mathcal{L}_{\text {match}}\left(y_{i}, \hat{y}_{\tau(i)}\right),
\label{eq:matching}
\end{equation}
where $\mathcal{L}_{\text {match}}\left(y_{i}, \hat{y}_{\tau(i)}\right)$ is the pair-wise loss between ground truth and the prediction with index $\tau(i)$.

\subsection{Model Improvement}
In this section, we make slight adjustments to data-hungry detection transformers such as DETR and CondDETR, to largely boost their data efficiency.

\noindent\textbf{Sparse Feature Sampling.}
\begin{figure*}[t]
  \centerline{\includegraphics[width=\textwidth]{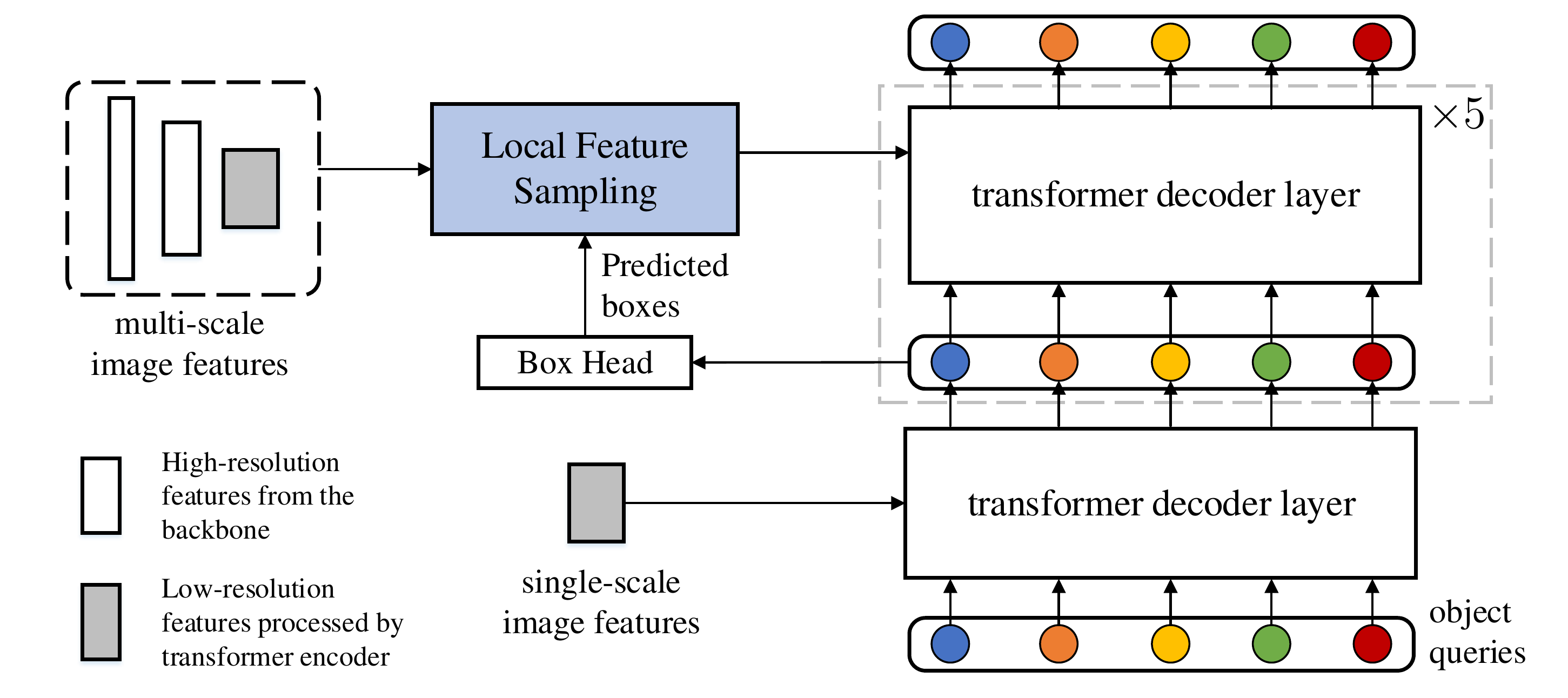}}
  \caption{The proposed data-efficient detection transformer structure. With minimum modifications, we perform sparse sampling feature on key and value feature sent to the cross-attention layers in the decoder, under the guidance of bounding boxes predicted by the previous layer. Note the box head is part of the original detection transformers, which utilize deep supervision on the predictions made by each decoder layer. The backbone, the transformer encoder, and the first decoder layer are kept unchanged. 
}
\label{fig:model}
\end{figure*}
From the analysis in Section~\ref{sec:model_transform}, we can see that local feature sampling is critical to data efficiency. Fortunately, in detection transformers, the object locations are predicted after each decoder layer. Therefore, we can sample local features under the guidance of the bounding box prediction made by the previous decoder layer without introducing new parameters, as shown in Fig.~\ref{fig:model}. Although more sophisticated local feature sampling methods can be used, we simply adopt the commonly used RoIAlign~\cite{mask-rcnn}. Formally, the sampling operation can be written as:
\begin{equation}
\label{eq:sample_ss}
    \mathbf{z}_{\ell}^{L}=\text{RoIAlign}\left(z_{e}^{L}, \mathbf{b}_{\ell-1}\right), \quad \ell=2 \ldots L_{\mathrm{d}}
\end{equation}
where $\mathbf{b}_{\ell-1}$ is the bounding boxes predicted by the previous layer, $\mathbf{z}_{\ell}^{L} \in \mathbb{R}^{N \times {K}^{2} \times D}$ is the sampled feature, $K$ is the feature resolution in RoIAlign sampling. Note the reshape and flatten operations are omitted in Equation~\ref{eq:sample_ss}. Similarly, the corresponding positional embedding $p_{\ell}^{L}$ can be obtained. 

The cascaded structure in the detection transformer makes it natural to use layer-wise bounding box refinement~\cite{cascade-rcnn,zhu2020deformable} to improve detection performance. Our experiments in Section~\ref{sec:model_transform} also validate the effectiveness of the iterative refinement and making predictions with respect to initial spatial references. For this reason, we also introduce bounding box refinement and initial reference points during our implementation, as did in CondDETR~\cite{meng2021conditional}.

\noindent\textbf{Incorporating Multi-scale Feature.}
Our sparse feature sampling makes it possible to use multi-scale features in detection transformers with little computation cost.
To this end, we also flatten and embed the high-resolution features extracted by the backbone to obtain $\left\{z^{l}\right\}_{l=1}^{L-1} \in \mathbb{R}^{S^{l} \times D}$ for local feature sampling. However, these features are not processed by the transformer encoder. Although more sophisticated techniques can be used, these single-scale features sampled by RoIAlign are simply concatenated to form our multi-scale feature. These features are naturally fused by the cross-attention in the decoder.
\begin{equation}
    \mathbf{z}_{\ell}^{\mathrm{ms}}=\left[\mathbf{z}_{\ell}^{1}, \mathbf{z}_{\ell}^{2}, \ldots, \mathbf{z}_{\ell}^{L}\right], \ell=2 \ldots L_{\mathrm{d}},
\end{equation}
where $\mathbf{z}_{\ell}^{\mathrm{ms}} \in \mathbb{R}^{N \times L{K}^{2} \times D}$ is the multi-scale feature, and $\mathbf{z}_{\ell}^{l} = \operatorname{RoIAlign}\left(z^{l}, \mathbf{b}_{\ell-1}\right), l=1 \ldots L-1$. The corresponding positional embedding $\mathbf{p}_{\ell}^{\mathrm{ms}}$ is obtained in a similar way. The decoding process is the same as original detection transformers, as shown in Equation~\ref{eq:decoder}, where we have $\mathbf{z}_{\ell}=\mathbf{z}_{\ell}^{\mathrm{ms}}$ and ${p}_{\ell}={p}_{\ell}^{\mathrm{ms}}$. Please refer to the Appendix for details in implementation.

\subsection{Label Augmentation for Richer Supervision}
\begin{figure*}[t]
  \centerline{\includegraphics[width=\textwidth]{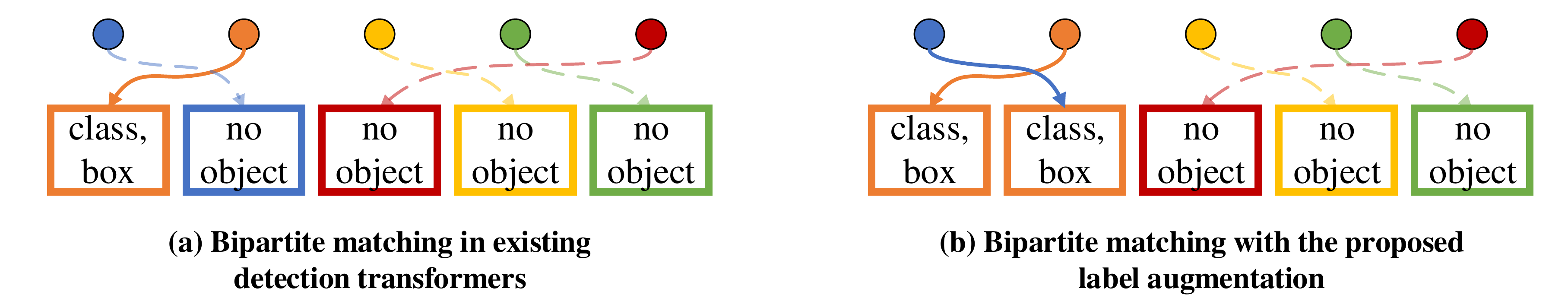}}
  \caption{Illustration of the proposed label augmentation method. 
  The predictions and the ground truths are represented by circles and rectangles, respectively.
  The matching between foreground instances is represented by solid lines, while the matching between background instances is represented by dotted lines. The prediction in blue that was originally matched to a background instance in (a) is now matched to a foreground instance in our method (b), thus obtaining more abundant supervision.}
  \label{fig:label_repeat}
\end{figure*}

Detection transformers perform one-to-one matching for label assignment, 
which means only a small number of detection candidates are provided with a positive supervision signal in each iteration. As a result, the model has to obtain enough supervision from a larger amount of data or more training epochs.

To alleviate this problem, we propose a label augmentation strategy to provide a richer supervised signal to the detection transformers, by simply repeating positive labels during bipartite matching. As shown in Fig.~\ref{fig:label_repeat}, we repeat the labels of each foreground sample $y_i$ for $R_i$ times, while keeping the total length of the label set $N$ unchanged.
\begin{equation}
    y=\left\{y_{1}^{1}, y_{1}^{2}, \ldots, y_{1}^{R_{1}}, \ldots, y_{M}^{1}, y_{M}^{2}, \ldots, y_{M}^{R_{M}}, \ldots, \varnothing, \ldots, \varnothing\right\}.
\end{equation}
Subsequently, the label assignment is achieved according to the operation in Equation~\ref{eq:matching}. 

Two label repeat strategies are considered during our implementation as follows. (a) Fixed repeat times, where all positive labels are repeated for the same number of times, \ie, $R_{i} = R, i=1 \ldots M$. (b) Fixed positive sample ratio, where the positive labels are sampled repeatedly to ensure a proportion of $r$ positive samples in the label set. Specifically, $F = N \times r$ is the expected number of positive samples after repeating labels. We first repeat each positive label for $F // M$ times, and subsequently, randomly sample $F \% M$ positive labels without repetition.
By default, we use the fixed repeat times strategy, because it is easier to implement and the resultant label set is deterministic.

\section{Experiments}
\label{sec:experiments}
\noindent\textbf{Datasets.}
To explore detection transformers' data efficiency, most of our experiments are conducted on small-size datasets including Cityscapes~\cite{cityscapes} and sub-sampled COCO 2017~\cite{coco}. Cityscapes contains 2,975 images for training and 500 images for evaluation. For the sub-sampled COCO 2017 dataset, the training images are randomly sub-sampled by 0.1, 0.05, 0.02, and 0.01, while the evaluation set is kept unchanged. Besides, we also validate the effectiveness of our method on the full-size COCO 2017 dataset with 118K training images.

\noindent\textbf{Implementation details.}
By default, our feature sampling is implemented as RoIAlign with a feature resolution of 4. Three different feature levels are included for multi-scale feature fusion. A fixed repeat time of 2 is adopted for our label augmentation and non-maximum suppression (NMS) with a threshold of 0.7 is used for duplicate removal. All models are trained for 50 epochs and the learning rate decays after 40 epochs, unless specified.
ResNet-50~\cite{resnet} pre-trained on ImageNet-1K~\cite{imagenet} is used as backbone. To guarantee enough number of training iterations, all experiments on Cityscapes and sub-sampled COCO 2017 datasets are trained with a batch size of 8. And the results are averaged over five repeated runs with different random seeds. Our data-efficient detection transformers only make slight modifications to existing methods. Unless specified, we follow the original implementation details of corresponding baseline methods~\cite{detr,meng2021conditional}. Run time is evaluated on NVIDIA A100 GPU.

\subsection{Main Results}

\begin{table}[ht]
\renewcommand{\arraystretch}{1.0}
\renewcommand{\tabcolsep}{3.pt}
\centering
\caption{Comparison of detection transformers on Cityscapes. DE denotes data-efficient and LA denotes label augmentation. $\dagger$ indicates the query number is increased from 100 to 300.
}
\label{table: compare_city}
\resizebox{\textwidth}{!}{
    \begin{tabular}{l|c|cccccc|ccc}
    \Xhline{2\arrayrulewidth}  
    \hline
    \rowcolor{mygray}
    Method & Epochs & AP & AP$_\text{50}$ & AP$_\text{75}$ & AP$_\text{S}$ & AP$_\text{M}$ & AP$_\text{L}$ & \tabincell{c}{Params} & \tabincell{c}{FLOPs} & FPS \\
    \hline
    DETR~\cite{detr} & 300 & 11.5 & 26.7 & 8.6 & 2.5 & 9.5 & 25.1 & 41M & 86G & 44 \\
    UP-DETR~\cite{up-detr} & 300 & 23.8 & 45.7 & 20.8 & 4.0 & 20.3 & 46.6 & 41M & 86G & 44 \\
    PnP-DETR-$\alpha$=0.33~\cite{pnp-detr} & 300 & 11.2 & 11.5 & 8.7 & 2.3 & 21.2 & 25.6 & 41M & 79G & 43 \\
    PnP-DETR-$\alpha$=0.80~\cite{pnp-detr} & 300 & 11.4 & 26.6 & 8.1 & 2.5 & 9.3 & 24.7 & 41M & 83G & 43 \\
    CondDETR~\cite{meng2021conditional} & 50 & 12.1 & 28.0 & 9.1 & 2.2 & 9.8 & 27.0 & 43M & 90G & 39 \\
    SMCA (single scale)~\cite{smca} & 50 & 14.7 & 32.9 & 11.6 & 2.9 & 12.9 & 30.9 & 42M & 86G & 39 \\
    DeformDETR~\cite{zhu2020deformable} & 50 & 27.3 & 49.2 & 26.3 & 8.7 & 28.2 & 45.7 & 40M & 174G & 28 \\
    \hline
    DE-DETR & 50 & 21.7 & 41.7 & 19.2 & 4.9 & 20.0 & 39.9 & 42M & 88G & 34 \\ 
    DELA-DETR$^\dagger$ & 50 & 24.5 & 46.2 & 22.5 & 6.1 & 23.3 & 43.9 & 42M & 91G & 29 \\  
    \hline
    DE-CondDETR & 50 & 26.8 & 47.8 & 25.4 & 6.8 & 25.6 & 46.6 & 44M & 107G & 29 \\
    DELA-CondDETR & 50 & 29.5 & 52.8 & 27.6 & 7.5 & 28.2 & 50.1 & 44M & 107G & 29 \\
    
    %
    \Xhline{2\arrayrulewidth}
    \end{tabular}
}
\end{table}

\noindent\textbf{Results on Cityscapes.}
In this section, we compare our method with existing detection transformers. As shown in Table~\ref{table: compare_city}, most of them suffer from the data-efficiency issue. Nevertheless, with minor changes to the CondDETR model, our DE-CondDETR is able to achieve comparable data efficiency to DeformDETR. Further, with the richer supervision provided by label augmentation, our DELA-CondDETR surpasses DeformDETR by 2.2 AP. Besides, our method can be combined with other detection transformers to significantly improve their data efficiency, for example, our DE-DETR and DELA-DETR trained for 50 epochs perform significantly better than DETR trained for 500 epochs. 

\begin{figure*}[t]
  \centerline{\includegraphics[width=\textwidth]{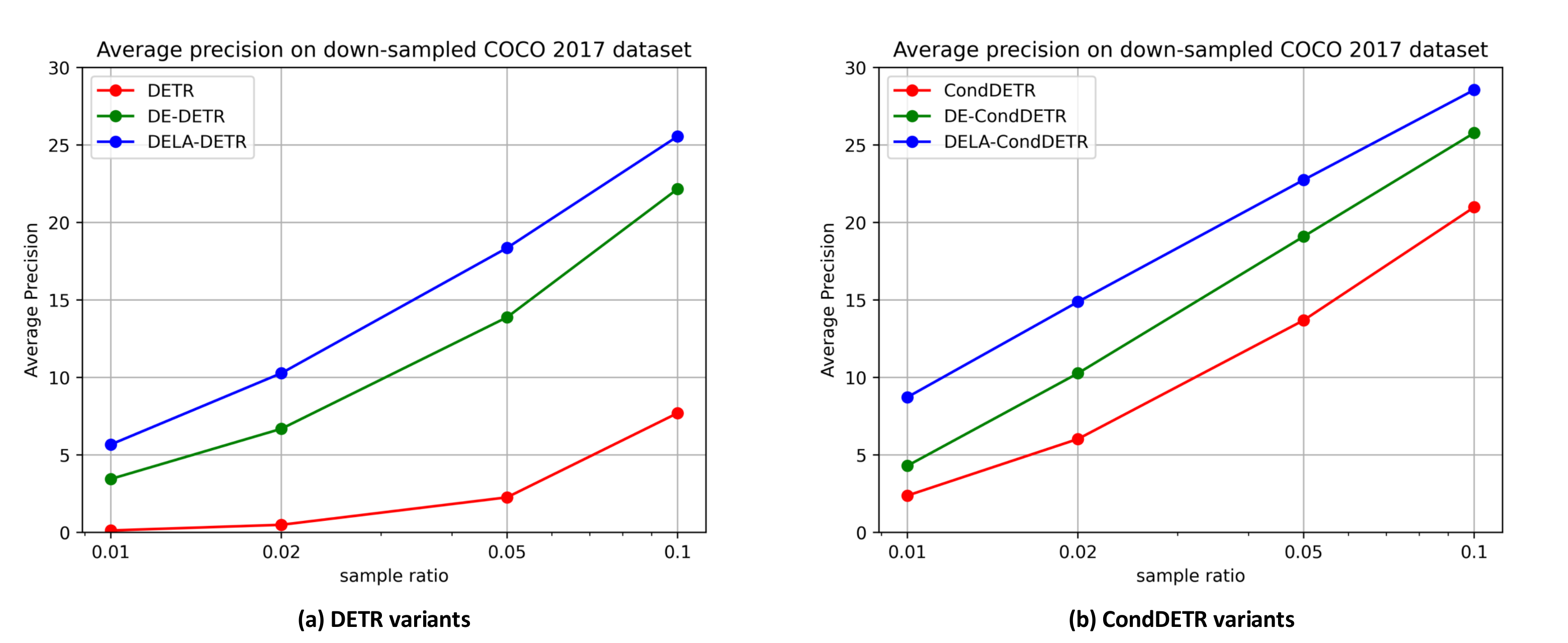}}
  \caption{Performance comparison of different methods on sub-sampled COCO 2017 dataset. Note the sample ratio is shown on a logarithmic scale. As can be seen, both local feature sampling and label augmentation consistently improve the model performance under varying data sampling ratios.}
  \label{fig:cocodown}
\end{figure*}

\noindent\textbf{Results on sub-sampled COCO 2017.}
Sub-sampled COCO 2017 datasets contain 11,828 (10\%), 5,914 (5\%), 2,365 (2\%), and 1,182 (1\%) training images, respectively. As shown in Fig~\ref{fig:cocodown}, our method consistently outperforms the baseline methods by a large margin. In particular, DELA-DETR trained with only $\sim$1K images significantly outperforms the DETR baseline with five times the training data. Similarly, DELA-CondDETR consistently outperforms the CondDETR baseline trained with twice the data volume.

\subsection{Ablations}
\label{subsec:ablations}
In this section, we perform ablated experiments to better understand each component of our method. All the ablation studies are implemented on the DELA-CondDETR and the Cityscapes dataset, while more ablation studies based on DELA-DETR can be found in our Appendix.

\begin{table}[ht]
\renewcommand{\arraystretch}{1.0}
\renewcommand{\tabcolsep}{2.pt}
\centering
\caption{Ablations on each component in DELA-CondDETR. ``SF", ``MS", and ``LA" represent sparse feature sampling, multi-scale feature fusion, and label augmentation.}
\label{table: ablate_main}
\resizebox{\textwidth}{!}{
    \begin{tabular}{l|ccc|cccccc|ccc}
    \Xhline{2\arrayrulewidth}  
    \hline
    \rowcolor{mygray}
    Method & SF & MS & LA & AP & AP$_\text{50}$ & AP$_\text{75}$ & AP$_\text{S}$ & AP$_\text{M}$ & AP$_\text{L}$ & \tabincell{c}{Params} & \tabincell{c}{FLOPs} & FPS \\
    \hline
    CondDETR~\cite{meng2021conditional} &  &  &  & 12.1 & 28.0 & 9.1 & 2.2 & 9.8 & 27.0 & 43M & 90G & 39 \\
    \hline
    & &  & $\checkmark$ & 14.7 & 31.6 & 12.1 & 2.9 & 12.5 & 32.1 & 43M & 90G & 38 \\
    & $\checkmark$ &  &  & 20.4 & 40.7 & 17.7 & 2.9 & 16.9 & 42.0 & 44M & 95G & 32 \\
    \hline 
    DE-CondDETR & $\checkmark$ & $\checkmark$ &  & 26.8 & 47.8 & 25.4 & 6.8 & 25.6 & 46.6 & 44M & 107G & 29 \\
    DELA-CondDETR & $\checkmark$ & $\checkmark$ & $\checkmark$ & 29.5 & 52.8 & 27.6 & 7.5 & 28.2 & 50.1 & 44M & 107G & 29 \\
    \Xhline{2\arrayrulewidth}
    \end{tabular}
}
\end{table}

\noindent\textbf{Effectiveness of each module.}
We first ablate the role of each module in our method, as shown in Table~\ref{table: ablate_main}. The use of local feature sampling and multi-scale feature fusion significantly improves the performance of the model by 8.3 and 6.4 AP, respectively. In addition, label augmentation further improves the performance by 2.7 AP. Besides, using it alone also brings a gain of 2.6 AP.

\begin{table}[ht]
\renewcommand{\arraystretch}{1.0}
\renewcommand{\tabcolsep}{2.pt}
\centering
\caption{Ablations on multi-scale feature levels and feature resolutions for RoIAlign. Note label augmentation is not utilized for clarity.}
\label{table: ablate_ms_res}
    \begin{tabular}{cc|cccccc|ccc}
    \Xhline{2\arrayrulewidth}  
    \hline
    \rowcolor{mygray}
    MS Lvls & RoI Res. & AP & AP$_\text{50}$ & AP$_\text{75}$ & AP$_\text{S}$ & AP$_\text{M}$ & AP$_\text{L}$ & \tabincell{c}{Params} & \tabincell{c}{FLOPs} & FPS \\
    
    \hline
    1 & 1 & 14.8 & 35.1 & 11.0 & 2.4 & 11.7 & 31.1 & 44M & 90G & 32 \\
    1 & 4 & 20.4 & 40.7 & 17.7 & 2.9 & 16.9 & 42.0 & 44M & 95G & 32 \\
    1 & 7 & 20.7 & 40.9 & 18.5 & 2.9 & 16.8 & 42.7 & 44M & 104G & 31 \\
    \hline
    3 & 4 & 26.8 & 47.8 & 25.4 & 6.8 & 25.6 & 46.6 & 44M & 107G & 29 \\
    4 & 4 & 26.3 & 47.1 & 25.1 & 6.5 & 24.8 & 46.5 & 49M & 112G & 28 \\
    \Xhline{2\arrayrulewidth}
    \end{tabular} 
\end{table}

\noindent\textbf{Feature resolution for RoIAlign.}
In general, a larger sample resolution in RoIAlign provides richer information and thus improves detection performance. However, sampling larger feature resolution is also more time-consuming and increases the computational cost of the decoding process. As shown in Table~\ref{table: ablate_ms_res}, the model performance is significantly improved by 5.6 AP when the resolution is increased from 1 to 4. However, when the resolution is further increased to 7, the improvement is minor and the FLOPs and latency are increased. For this reason, we set the feature resolution for RoIAlign as 4 by default. 

\noindent\textbf{Number of multi-scale features.}
To incorporate multi-scale features, we also sample the 8$\times$ and 16$\times$ down-sampled features from the backbone to construct multi-scale features of 3 different levels. As can be seen from Table~\ref{table: ablate_ms_res}, it significantly improves the model performance by 6.4 AP. However, when we further add the 64$\times$ down-sampled features for multi-scale fusion, the performance drops by 0.5 AP. 
By default, we use 3 feature levels for multi-scale feature fusion.


\begin{table}[tb]
	\begin{minipage}[t]{0.475\linewidth}
			\caption{
				\small{Ablations on label augmentation using fixed repeat time. 
				Params, FLOPs, and FPS are omitted since they are consistent for all settings.
				}
			}
			\label{tab:fix_time}
                \begin{tabular}{c|cccccc}
                \Xhline{2\arrayrulewidth}  
                \hline
                \rowcolor{mygray}
                Time & AP & AP$_\text{50}$ & AP$_\text{75}$ & AP$_\text{S}$ & AP$_\text{M}$ & AP$_\text{L}$ \\
                \hline
                -- & 26.8 & 47.8 & 25.4 & 6.8 & 25.6 & 46.6 \\
                \hline
                2 & 29.5 & 52.8 & 27.6 & 7.5 & 28.2 & 50.1 \\
                3 & 29.4 & 52.6 & 28.0 & 7.6 & 28.1 & 50.3 \\
                4 & 29.0 & 52.0 & 27.7 & 7.8 & 27.9 & 49.5 \\
                5 & 28.7 & 51.3 & 27.4 & 7.8 & 27.7 & 49.3 \\
                \Xhline{2\arrayrulewidth}
                \end{tabular}
	\end{minipage}
\hfill
	\begin{minipage}[t]{0.475\linewidth}
			\caption{
				\small{Ablations on label augmentation using fixed positive sample ratio.}
			}
			\label{tab:fix_ratio}
			\begin{center}
                \begin{tabular}{c|cccccc}
                \Xhline{2\arrayrulewidth}  
                \hline
                \rowcolor{mygray}
                Ratio & AP & AP$_\text{50}$ & AP$_\text{75}$ & AP$_\text{S}$ & AP$_\text{M}$ & AP$_\text{L}$ \\
                \hline
                -- & 26.8 & 47.8 & 25.4 & 6.8 & 25.6 & 46.6 \\
                \hline
                0.1 & 27.7 & 49.7 & 26.1 & 7.4 & 26.5 & 47.2 \\
                0.2 & 28.2 & 50.2 & 26.9 & 7.4 & 26.8 & 48.5 \\
                0.25 & 28.3 & 50.5 & 27.2 & 7.5 & 27.1 & 48.3 \\
                0.3 & 27.9 & 50.3 & 26.5 & 7.3 & 27.1 & 47.4 \\
                0.4 & 27.6 & 49.7 & 26.0 & 7.0 & 27.0 & 46.8 \\
                \Xhline{2\arrayrulewidth}
                \end{tabular}
			\end{center}
	\end{minipage}
\end{table}
\noindent\textbf{Strategies for label augmentation.}
In this section, we ablate the proposed two label augmentation strategies, namely fixed repeat time and fixed positive sample ratio. As shown in Table~\ref{tab:fix_time}, using different fixed repeated times consistently improves the performance of DE-DETR baseline, but the performance gain tends to decrease as the number of repetitions increases. 
Moreover, as shown in Table~\ref{tab:fix_ratio}, although using different ratios can bring improvement on AP, the best performance is achieved when the positive to negative samples ratio is 1:3, which, interestingly, is also the most commonly used positive to negative sampling ratio in the RCNN series detectors, \eg Faster RCNN.

\begin{table}[ht]
\renewcommand{\arraystretch}{1.0}
\renewcommand{\tabcolsep}{2.pt}
\centering
\caption{Performance of our data-efficient detection transformers on COCO 2017. All models are trained for 50 epochs.}
\label{table: coco}
    \begin{tabular}{l|c|cccccc|ccc}
    \Xhline{2\arrayrulewidth}  
    \hline
    \rowcolor{mygray}
    Method & Epochs & AP & AP$_\text{50}$ & AP$_\text{75}$ & AP$_\text{S}$ & AP$_\text{M}$ & AP$_\text{L}$ & \tabincell{c}{Params} & \tabincell{c}{FLOPs} & FPS \\
    \hline
    DETR~\cite{detr} & 50 & 33.6 & 54.6 & 34.2 & 13.2 & 35.7 & 53.5 & 41M & 86G & 43 \\
    DE-DETR & 50 & 40.2 & 60.4 & 43.2 & 23.3 & 42.1 & 56.4 & 43M & 88G & 33 \\
    DELA-DETR$^\dagger$ & 50 & 41.9 & 62.6 & 44.8 & 24.9 & 44.9 & 56.8 & 43M & 91G & 29 \\
    \hline
    CondDETR~\cite{meng2021conditional} & 50 & 40.2 & 61.1 & 42.6 & 19.9 & 43.6 & 58.7 & 43M & 90G & 39 \\
    DE-CondDETR & 50 & 41.7 & 62.4 & 44.9 & 24.4 & 44.5 & 56.3 & 44M & 107G & 28 \\
    DELA-CondDETR & 50 & 43.0 & 64.0 & 46.4 & 26.0 & 45.5 & 57.7 & 44M & 107G & 28 \\
    \Xhline{2\arrayrulewidth}
    \end{tabular}
\end{table}
\subsection{Generalization to Sample-Rich Dataset}
Although the above experiments show that our method can improve model performance when only limited training data is available, there is no guarantee that our method remains effective when the training data is sufficient. To this end, we evaluate our method on COCO 2017 with a sufficient amount of data. As can be seen from Table~\ref{table: coco}, our method does not degrade the model performance on COCO 2017. Conversely, it delivers a promising improvement. Specifically, DELA-DETR and DELA-CondDETR improve their corresponding baseline by 8.3 and 2.8 AP, respectively.

\section{Conclusion}
\label{sec:conclution}
In this paper, we identify the data-efficiency issue of detection transformers. Through step-by-step model transformation from Sparse RCNN to DETR, we find that sparse feature sampling from local areas holds the key to data efficiency. Based on these, we improve existing detection transformers by simply sampling multi-scale features under the guidance of predicted bounding boxes, with minimum modifications to the original models. In addition, we propose a simple yet effective label augmentation strategy to provide richer supervision and thus further alleviate the data-efficiency issue. Extensive experiments validate the effectiveness of our method. As transformers become increasingly popular for visual tasks, we hope our work will inspire the community to explore the data efficiency of transformers for different tasks.

\noindent\textbf{Acknowledgement.}
This work is supported by National Key R\&D Program of China under Grant 2020AAA0105701, National Natural Science Foundation of China (NSFC) under Grants 61872327, Major Special Science and Technology Project of Anhui (No. 012223665049), and the ARC project FL-170100117.

\clearpage
%
%
\bibliographystyle{splncs04}
\bibliography{egbib}

\begin{thebibliography}{10}
\providecommand{\url}[1]{\texttt{#1}}
\providecommand{\urlprefix}{URL }
\providecommand{\doi}[1]{https://doi.org/#1}

\bibitem{soft-nms}
Bodla, N., Singh, B., Chellappa, R., Davis, L.S.: Soft-nms--improving object
  detection with one line of code. In: Proceedings of the IEEE international
  conference on computer vision. pp. 5561--5569 (2017)

\bibitem{cascade-rcnn}
Cai, Z., Vasconcelos, N.: Cascade r-cnn: Delving into high quality object
  detection. In: Proceedings of the IEEE conference on computer vision and
  pattern recognition. pp. 6154--6162 (2018)

\bibitem{cao2022training}
Cao, Y.H., Yu, H., Wu, J.: Training vision transformers with only 2040 images.
  arXiv preprint arXiv:2201.10728  (2022)

\bibitem{detr}
Carion, N., Massa, F., Synnaeve, G., Usunier, N., Kirillov, A., Zagoruyko, S.:
  End-to-end object detection with transformers. In: European Conference on
  Computer Vision. pp. 213--229. Springer (2020)

\bibitem{chen2022recurrent}
Chen, Z., Zhang, J., Tao, D.: Recurrent glimpse-based decoder for detection
  with transformer. In: Proceedings of the IEEE/CVF Conference on Computer
  Vision and Pattern Recognition. pp. 5260--5269 (2022)

\bibitem{chen2021visformer}
Chen, Z., Xie, L., Niu, J., Liu, X., Wei, L., Tian, Q.: Visformer: The
  vision-friendly transformer. In: Proceedings of the IEEE/CVF International
  Conference on Computer Vision. pp. 589--598 (2021)

\bibitem{chu2021twins}
Chu, X., Tian, Z., Wang, Y., Zhang, B., Ren, H., Wei, X., Xia, H., Shen, C.:
  Twins: Revisiting the design of spatial attention in vision transformers.
  Advances in Neural Information Processing Systems  \textbf{34} (2021)

\bibitem{cityscapes}
Cordts, M., Omran, M., Ramos, S., Rehfeld, T., Enzweiler, M., Benenson, R.,
  Franke, U., Roth, S., Schiele, B.: The cityscapes dataset for semantic urban
  scene understanding. In: Proceedings of the IEEE conference on computer
  vision and pattern recognition. pp. 3213--3223 (2016)

\bibitem{up-detr}
Dai, Z., Cai, B., Lin, Y., Chen, J.: Up-detr: Unsupervised pre-training for
  object detection with transformers. In: Proceedings of the IEEE conference on
  computer vision and pattern recognition (2020)

\bibitem{imagenet}
Deng, J., Dong, W., Socher, R., Li, L.J., Li, K., Fei-Fei, L.: Imagenet: A
  large-scale hierarchical image database. In: 2009 IEEE conference on computer
  vision and pattern recognition. pp. 248--255. IEEE (2009)

\bibitem{dosovitskiy2020image}
Dosovitskiy, A., Beyer, L., Kolesnikov, A., Weissenborn, D., Zhai, X.,
  Unterthiner, T., Dehghani, M., Minderer, M., Heigold, G., Gelly, S., et~al.:
  An image is worth 16x16 words: Transformers for image recognition at scale.
  arXiv preprint arXiv:2010.11929  (2020)

\bibitem{voc}
Everingham, M., Van~Gool, L., Williams, C.K., Winn, J., Zisserman, A.: The
  pascal visual object classes (voc) challenge. International journal of
  computer vision  \textbf{88}(2),  303--338 (2010)

\bibitem{fang2021msg}
Fang, J., Xie, L., Wang, X., Zhang, X., Liu, W., Tian, Q.: Msg-transformer:
  Exchanging local spatial information by manipulating messenger tokens. arXiv
  preprint arXiv:2105.15168  (2021)

\bibitem{yolos}
Fang, Y., Liao, B., Wang, X., Fang, J., Qi, J., Wu, R., Niu, J., Liu, W.: You
  only look at one sequence: Rethinking transformer in vision through object
  detection. arXiv preprint arXiv:2106.00666  (2021)

\bibitem{dpm}
Felzenszwalb, P.F., Girshick, R.B., McAllester, D., Ramanan, D.: Object
  detection with discriminatively trained part-based models. IEEE transactions
  on pattern analysis and machine intelligence  \textbf{32}(9),  1627--1645
  (2009)

\bibitem{smca}
Gao, P., Zheng, M., Wang, X., Dai, J., Li, H.: Fast convergence of detr with
  spatially modulated co-attention. In: Proceedings of the IEEE international
  conference on computer vision (2021)

\bibitem{ota}
Ge, Z., Liu, S., Li, Z., Yoshie, O., Sun, J.: Ota: Optimal transport assignment
  for object detection. In: Proceedings of the IEEE/CVF Conference on Computer
  Vision and Pattern Recognition. pp. 303--312 (2021)

\bibitem{rcnn}
Girshick, R., Donahue, J., Darrell, T., Malik, J.: Rich feature hierarchies for
  accurate object detection and semantic segmentation. In: Proceedings of the
  IEEE conference on computer vision and pattern recognition. pp. 580--587
  (2014)

\bibitem{tnt}
Han, K., Xiao, A., Wu, E., Guo, J., Xu, C., Wang, Y.: Transformer in
  transformer. Advances in Neural Information Processing Systems  \textbf{34}
  (2021)

\bibitem{mask-rcnn}
He, K., Gkioxari, G., Doll{\'a}r, P., Girshick, R.: Mask r-cnn. In: Proceedings
  of the IEEE international conference on computer vision. pp. 2961--2969
  (2017)

\bibitem{resnet}
He, K., Zhang, X., Ren, S., Sun, J.: Deep residual learning for image
  recognition. In: Proceedings of the IEEE conference on computer vision and
  pattern recognition. pp. 770--778 (2016)

\bibitem{henaff2020data}
Henaff, O.: Data-efficient image recognition with contrastive predictive
  coding. In: International Conference on Machine Learning. pp. 4182--4192.
  PMLR (2020)

\bibitem{hu2018relation}
Hu, H., Gu, J., Zhang, Z., Dai, J., Wei, Y.: Relation networks for object
  detection. In: Proceedings of the IEEE conference on computer vision and
  pattern recognition. pp. 3588--3597 (2018)

\bibitem{jia2016dynamic}
Jia, X., De~Brabandere, B., Tuytelaars, T., Gool, L.V.: Dynamic filter
  networks. Advances in neural information processing systems  \textbf{29}
  (2016)

\bibitem{lee2015deeply}
Lee, C.Y., Xie, S., Gallagher, P., Zhang, Z., Tu, Z.: Deeply-supervised nets.
  In: Artificial intelligence and statistics. pp. 562--570. PMLR (2015)

\bibitem{fpn}
Lin, T.Y., Doll{\'a}r, P., Girshick, R., He, K., Hariharan, B., Belongie, S.:
  Feature pyramid networks for object detection. In: Proceedings of the IEEE
  conference on computer vision and pattern recognition. pp. 2117--2125 (2017)

\bibitem{lin2017focal}
Lin, T.Y., Goyal, P., Girshick, R., He, K., Doll{\'a}r, P.: Focal loss for
  dense object detection. In: Proceedings of the IEEE international conference
  on computer vision. pp. 2980--2988 (2017)

\bibitem{coco}
Lin, T.Y., Maire, M., Belongie, S., Hays, J., Perona, P., Ramanan, D.,
  Doll{\'a}r, P., Zitnick, C.L.: Microsoft coco: Common objects in context. In:
  European conference on computer vision (2014)

\bibitem{wb-detr}
Liu, F., Wei, H., Zhao, W., Li, G., Peng, J., Li, Z.: Wb-detr:
  Transformer-based detector without backbone. In: Proceedings of the IEEE/CVF
  International Conference on Computer Vision. pp. 2979--2987 (2021)

\bibitem{liu2021rankdetnet}
Liu, J., Li, D., Zheng, R., Tian, L., Shan, Y.: Rankdetnet: Delving into
  ranking constraints for object detection. In: Proceedings of the IEEE/CVF
  Conference on Computer Vision and Pattern Recognition. pp. 264--273 (2021)

\bibitem{ssd}
Liu, W., Anguelov, D., Erhan, D., Szegedy, C., Reed, S., Fu, C.Y., Berg, A.C.:
  Ssd: Single shot multibox detector. In: European conference on computer
  vision. pp. 21--37. Springer (2016)

\bibitem{liu2021efficient}
Liu, Y., Sangineto, E., Bi, W., Sebe, N., Lepri, B., Nadai, M.: Efficient
  training of visual transformers with small datasets. Advances in Neural
  Information Processing Systems  \textbf{34} (2021)

\bibitem{swin}
Liu, Z., Lin, Y., Cao, Y., Hu, H., Wei, Y., Zhang, Z., Lin, S., Guo, B.: Swin
  transformer: Hierarchical vision transformer using shifted windows. arXiv
  preprint arXiv:2103.14030  (2021)

\bibitem{meng2021conditional}
Meng, D., Chen, X., Fan, Z., Zeng, G., Li, H., Yuan, Y., Sun, L., Wang, J.:
  Conditional detr for fast training convergence. In: Proceedings of the IEEE
  international conference on computer vision (2021)

\bibitem{misra2021end}
Misra, I., Girdhar, R., Joulin, A.: An end-to-end transformer model for 3d
  object detection. In: Proceedings of the IEEE/CVF International Conference on
  Computer Vision. pp. 2906--2917 (2021)

\bibitem{yolo}
Redmon, J., Divvala, S., Girshick, R., Farhadi, A.: You only look once:
  Unified, real-time object detection. In: Proceedings of the IEEE conference
  on computer vision and pattern recognition. pp. 779--788 (2016)

\bibitem{faster-rcnn}
Ren, S., He, K., Girshick, R., Sun, J.: Faster r-cnn: towards real-time object
  detection with region proposal networks. IEEE transactions on pattern
  analysis and machine intelligence  \textbf{39}(6),  1137--1149 (2016)

\bibitem{shen2022acl}
Shen, Y., Wang, X., Tan, Z., Xu, G., Xie, P., Huang, F., Lu, W., Zhuang, Y.:
  Parallel instance query network for named entity recognition. In: Proceedings
  of the 60th Annual Meeting of the Association for Computational Linguistics.
  Association for Computational Linguistics (2022),
  \url{https://arxiv.org/abs/2203.10545}

\bibitem{sparse-rcnn}
Sun, P., Zhang, R., Jiang, Y., Kong, T., Xu, C., Zhan, W., Tomizuka, M., Li,
  L., Yuan, Z., Wang, C., et~al.: Sparse r-cnn: End-to-end object detection
  with learnable proposals. In: Proceedings of the IEEE/CVF Conference on
  Computer Vision and Pattern Recognition. pp. 14454--14463 (2021)

\bibitem{thomas2016data}
Thomas, P., Brunskill, E.: Data-efficient off-policy policy evaluation for
  reinforcement learning. In: International Conference on Machine Learning. pp.
  2139--2148. PMLR (2016)

\bibitem{tian2020conditional}
Tian, Z., Shen, C., Chen, H.: Conditional convolutions for instance
  segmentation. In: European Conference on Computer Vision. pp. 282--298.
  Springer (2020)

\bibitem{tian2019fcos}
Tian, Z., Shen, C., Chen, H., He, T.: Fcos: Fully convolutional one-stage
  object detection. In: Proceedings of the IEEE/CVF International Conference on
  Computer Vision. pp. 9627--9636 (2019)

\bibitem{deit}
Touvron, H., Cord, M., Douze, M., Massa, F., Sablayrolles, A., J{\'e}gou, H.:
  Training data-efficient image transformers \& distillation through attention.
  In: International Conference on Machine Learning. pp. 10347--10357. PMLR
  (2021)

\bibitem{vaswani2017attention}
Vaswani, A., Shazeer, N., Parmar, N., Uszkoreit, J., Jones, L., Gomez, A.N.,
  Kaiser, L., Polosukhin, I.: Attention is all you need. In: Conference on
  Neural Information Processing Systems (2017)

\bibitem{GuidedAnchor}
Wang, J., Chen, K., Yang, S., Loy, C.C., Lin, D.: Region proposal by guided
  anchoring. In: Proceedings of the IEEE/CVF Conference on Computer Vision and
  Pattern Recognition. pp. 2965--2974 (2019)

\bibitem{pnp-detr}
Wang, T., Yuan, L., Chen, Y., Feng, J., Yan, S.: Pnp-detr: towards efficient
  visual analysis with transformers. In: Proceedings of the IEEE/CVF
  International Conference on Computer Vision (2021)

\bibitem{wang2022fpdetr}
Wang, W., Cao, Y., Zhang, J., Tao, D.: {FP}-{DETR}: Detection transformer
  advanced by fully pre-training. In: International Conference on Learning
  Representations (2022), \url{https://openreview.net/forum?id=yjMQuLLcGWK}

\bibitem{pvt}
Wang, W., Xie, E., Li, X., Fan, D.P., Song, K., Liang, D., Lu, T., Luo, P.,
  Shao, L.: Pyramid vision transformer: A versatile backbone for dense
  prediction without convolutions. In: Proceedings of the IEEE/CVF
  International Conference on Computer Vision. pp. 568--578 (2021)

\bibitem{wang2021end}
Wang, Y., Xu, Z., Wang, X., Shen, C., Cheng, B., Shen, H., Xia, H.: End-to-end
  video instance segmentation with transformers. In: Proceedings of the
  IEEE/CVF Conference on Computer Vision and Pattern Recognition. pp.
  8741--8750 (2021)

\bibitem{xie2021segformer}
Xie, E., Wang, W., Yu, Z., Anandkumar, A., Alvarez, J.M., Luo, P.: Segformer:
  Simple and efficient design for semantic segmentation with transformers.
  Advances in Neural Information Processing Systems  \textbf{34} (2021)

\bibitem{xu2021vitae}
Xu, Y., Zhang, Q., Zhang, J., Tao, D.: Vitae: Vision transformer advanced by
  exploring intrinsic inductive bias. Advances in Neural Information Processing
  Systems  \textbf{34} (2021)

\bibitem{MetaAnchor}
Yang, T., Zhang, X., Li, Z., Zhang, W., Sun, J.: Metaanchor: Learning to detect
  objects with customized anchors. Advances in neural information processing
  systems  \textbf{31} (2018)

\bibitem{yuan2021polyphonicformer}
Yuan, H., Li, X., Yang, Y., Cheng, G., Zhang, J., Tong, Y., Zhang, L., Tao, D.:
  Polyphonicformer: Unified query learning for depth-aware video panoptic
  segmentation. In: European Conference on Computer Vision (2022)

\bibitem{t2t}
Yuan, L., Chen, Y., Wang, T., Yu, W., Shi, Y., Jiang, Z.H., Tay, F.E., Feng,
  J., Yan, S.: Tokens-to-token vit: Training vision transformers from scratch
  on imagenet. In: Proceedings of the IEEE/CVF International Conference on
  Computer Vision. pp. 558--567 (2021)

\bibitem{zhang2022vitaev2}
Zhang, Q., Xu, Y., Zhang, J., Tao, D.: Vitaev2: Vision transformer advanced by
  exploring inductive bias for image recognition and beyond. arXiv preprint
  arXiv:2202.10108  (2022)

\bibitem{atss}
Zhang, S., Chi, C., Yao, Y., Lei, Z., Li, S.Z.: Bridging the gap between
  anchor-based and anchor-free detection via adaptive training sample
  selection. In: Proceedings of the IEEE/CVF conference on computer vision and
  pattern recognition. pp. 9759--9768 (2020)

\bibitem{FreeAnchor}
Zhang, X., Wan, F., Liu, C., Ji, R., Ye, Q.: Freeanchor: Learning to match
  anchors for visual object detection. Advances in neural information
  processing systems  \textbf{32} (2019)

\bibitem{zheng2021rethinking}
Zheng, S., Lu, J., Zhao, H., Zhu, X., Luo, Z., Wang, Y., Fu, Y., Feng, J.,
  Xiang, T., Torr, P.H., et~al.: Rethinking semantic segmentation from a
  sequence-to-sequence perspective with transformers. In: Proceedings of the
  IEEE/CVF conference on computer vision and pattern recognition. pp.
  6881--6890 (2021)

\bibitem{autoassign}
Zhu, B., Wang, J., Jiang, Z., Zong, F., Liu, S., Li, Z., Sun, J.: Autoassign:
  Differentiable label assignment for dense object detection. arXiv preprint
  arXiv:2007.03496  (2020)

\bibitem{zhu2020deformable}
Zhu, X., Su, W., Lu, L., Li, B., Wang, X., Dai, J.: Deformable detr: Deformable
  transformers for end-to-end object detection. In: International Conference on
  Learning and Representations (2020)

\end{thebibliography}

\clearpage

\appendix

\section{Data Efficiency}
Although there is no strict definition of data efficiency, it has been studied under various contexts~\cite{deit,liu2021efficient,thomas2016data,henaff2020data}. As described in Section 1 of our main text, the requirement of more training data brings two issues: (a) more human labors are needed to collect and label enough training data; (b) more computational costs are required to train the model. In this paper, we aim to alleviate both these two issues of existing detection transformers. Thus, most of our experiments are performed on small-size datasets and implemented with a short training schedule of 50 epochs. 

\begin{table}[ht]
\renewcommand{\arraystretch}{1.0}
\renewcommand{\tabcolsep}{2.pt}
\centering
\caption{Performance comparison of CondDETR variants trained with a long training schedule, experimented on down-sampled COCO 2017 dataset. ``Rate" indicates the sample rate. The number of training epochs is increased to ensure the computational cost is the same as that of full COCO 2017 training.}
\label{table: app_cocodown_long}
\resizebox{\textwidth}{!}{
    \begin{tabular}{c|c|c|cccccc|ccc}
    \Xhline{2\arrayrulewidth}  
    \hline
    \rowcolor{mygray}
    Rate & Method & Epochs & AP & AP$_\text{50}$ & AP$_\text{75}$ & AP$_\text{S}$ & AP$_\text{M}$ & AP$_\text{L}$ & \tabincell{c}{Params} & \tabincell{c}{FLOPs} & FPS \\
    \hline
    \multirow{3}{*}{0.01} & CondDETR~\cite{meng2021conditional} & 5000 & 10.0 & 21.1 & 8.3 & 2.6 & 10.7 & 15.7 & 43M & 90G & 39 \\
    & DE-CondDETR & 5000 & 13.2 & 23.5 & 12.8 & 5.4 & 14.2 & 17.3 & 44M & 107G & 28 \\
    & DELA-CondDETR & 5000 & 13.4 & 24.1 & 13.0 & 5.6 & 14.4 & 17.7 & 44M & 107G & 28 \\
    \hline
    \multirow{3}{*}{0.02} & CondDETR~\cite{meng2021conditional} & 2500 & 14.7 & 28.6 & 13.3 & 4.3 & 14.8 & 23.7 & 43M & 90G & 39 \\
    & DE-CondDETR & 2500 & 17.5 & 30.7 & 17.4 & 7.3 & 18.9 & 24.7 & 44M & 107G & 28 \\
    & DELA-CondDETR & 2500 & 18.4 & 32.0 & 18.6 & 8.0 & 18.8 & 25.9 & 44M & 107G & 28 \\
    \hline
    \multirow{3}{*}{0.05} & CondDETR~\cite{meng2021conditional} & 1000 & 20.1 & 36.8 & 19.4 & 6.5 & 21.0 & 30.9 & 43M & 90G & 39 \\
    & DE-CondDETR & 1000 & 23.3 & 39.3 & 23.7 & 10.5 & 24.9 & 32.4 & 44M & 107G & 28 \\
    & DELA-CondDETR & 1000 & 23.8 & 40.0 & 24.4 & 10.7 & 25.4 & 33.2 & 44M & 107G & 28 \\
    \hline
    \multirow{3}{*}{0.1} & CondDETR~\cite{meng2021conditional} & 500 & 24.9 & 43.6 & 24.4 & 8.9 & 26.6 & 37.7 & 43M & 90G & 39 \\
    & DE-CondDETR & 500 & 27.5 & 45.0 & 28.2 & 13.6 & 29.1 & 37.0 & 44M & 107G & 28 \\
    & DELA-CondDETR & 500 & 28.3 & 46.1 & 29.4 & 14.4 & 30.0 & 38.5 & 44M & 107G & 28 \\
    \hline
    \Xhline{2\arrayrulewidth}
    \end{tabular}
}
\end{table}

In this section, we ablate the effectiveness of our method on alleviating human labor, by training on small-size datasets but with a much longer training schedule. Specifically, we use the same sub-sampled COCO 2017 dataset in Section 5 of our main text. But we increase the number of training epochs to match the computational cost of full COCO 2017 dataset training. The experiments are based on CondDETR~\cite{meng2021conditional} variants. All models are trained with a batch size of 32 and the learning rate is decayed after training for 0.8 times the total training epochs.

As can be seen from Table~\ref{table: app_cocodown_long}, with a much longer training schedule and more training costs, the performance of the CondDETR baseline is improved compared with the results under a short training schedule in Fig. 4 (b) of our manuscript, especially when the number of training images is small. However, our DE-CondDETR and DE-CondDETR still consistently outperform the CondDETR baseline by a large margin, which manifests the effectiveness of our method in alleviating human labor for data collection.

\section{Fine-tuning from COCO pre-trained model weights}
\begin{table}[ht]
\renewcommand{\arraystretch}{1.0}
\renewcommand{\tabcolsep}{2.pt}
\centering
\caption{Training on Cityscapes with COCO pre-trained weights.
}
\label{table: pretrained}
    \begin{tabular}{l|c|c|c|cccccc}
    \Xhline{2\arrayrulewidth}  
    \hline
    \rowcolor{mygray}
     & \multicolumn{2}{c|}{Pre-training on COCO} &  \multicolumn{4}{c}{Fine-tuning on Cityscapes} \\ 
    \cline{2-7} 
    \rowcolor{mygray}
    \multirow{-2}{*}{Method}
    & Epochs & AP & Epochs & AP & AP$_\text{50}$ & AP$_\text{75}$ \\
    
    \hline
    \multirow{2}{*}{DETR} & \multirow{2}{*}{500} & \multirow{2}{*}{42.0} & 50 & 25.7 & 47.6 & 24.3 \\ 
     & & & 300 & 29.4 & 51.5 & 27.7 \\ 
    \hline
    \multirow{2}{*}{\textbf{DELA}-DETR} & \multicolumn{2}{c|}{N/A (Scratch)} & 50 & 24.5 & 46.2 & 22.5  \\ 
    \cline{2-3} 
    \cline{2-3} 
     & 50 & 41.9 & 50 & 33.4 & 54.9 & 33.4  \\ 
    \Xhline{2\arrayrulewidth} 
    CondDETR & 50 & 40.2 & 50 & 29.8 & 55.1 & 27.5  \\ 
    \hline
    \multirow{2}{*}{\textbf{DELA}-CondDETR} & \multicolumn{2}{c|}{N/A (Scratch)} & 50 & 29.5 & 52.8 & 27.6 \\ 
    \cline{2-3} 
    \cline{2-3} 
     & 50 & 43.0 & 50 & 35.1 & 58.6 & 35.2  \\ 
    \Xhline{2\arrayrulewidth} 
    
    \end{tabular}
\end{table}
We conduct experiments to train detection transformers from COCO pre-trained weights. As shown in Table~\ref{table: pretrained}, the performance of DETR and CondDETR are significantly improved, and slightly outperform our DELA models trained from scratch. However, even better results can be achieved when our DELA models are trained from their corresponding COCO pre-trained weights. 

What's more interesting is that our DELA models can also benefit from COCO pre-trained DETR or CondDETR weights, thanks to our minimum modifications to the model structures. 
For example, DELA-CondDETR achieves 32.4 AP when fine-tuned from COCO pre-trained CondDETR, and DELA-DETR achieve 28.6 AP when fine-tuned from COCO pre-trained DETR.

\section{More Implementation Details}

\noindent\textbf{More details for experiments in Fig. 1.}
Experiment results on both the sample-rich COCO 2017 dataset and the small-size Cityscapes dataset are summarized in Fig. 1 of our manuscript. The results on COCO 2017 are collected from the corresponding original papers, while the results on Cityscapes are based on our re-implementations. Specifically, we follow the default training setting of corresponding methods~\cite{faster-rcnn,sparse-rcnn,zhu2020deformable,detr,meng2021conditional,pnp-detr,smca}. The only difference is that we use a small batch size of 8 to guarantee enough training iterations on the small-size dataset. For SMCA~\cite{smca}, since only the single-scale version is made publicly available, we adopt its single-scale variant, denoted as SMCA-SS. All Cityscapes experiments are repeated for 5 runs with different random seeds and the averaged results are reported.

\noindent\textbf{More details for the model transformation.}
Similar to other experiments on small-size datasets, we use a small batch size of 8 to guarantee enough training iterations on Cityscapes. Sparse RCNN is trained with focal loss~\cite{lin2017focal} and 300 queries, under the same data augmentation pipeline as DETR.

\noindent\textbf{More details for the experiments in Section 5.}
(a) For the ablation study on the number of multi-scale features, the 64$\times$ down-sampled feature is obtained by applying a 3$\times$3 convolution layer with a stride of 2 on the 32$\times$ down-sampled feature, following DeformDETR~\cite{zhu2020deformable}.
(b) For the evaluation of generalization to the sample-rich dataset, we re-implement DETR and CondDETR on COCO 2017 dataset for a fair comparison with our method.

\section{Pseudo-code for the implementation of DE-DETRs}
\lstset{
  backgroundcolor=\color{white},
  basicstyle=\fontsize{7.5pt}{8.5pt}\fontfamily{lmtt}\selectfont,
  columns=fullflexible,
  breaklines=true,
  captionpos=b,
  commentstyle=\fontsize{8pt}{9pt}\color{codegray},
  keywordstyle=\fontsize{8pt}{9pt}\color{codegreen},
  stringstyle=\fontsize{8pt}{9pt}\color{codeblue},
  frame=tb,
  otherkeywords = {self},
}
\begin{figure}[ht]
\begin{lstlisting}[language=python]
def forward(image_feats, query_feats, bbox):
    # image_feats: (B, D, H, W), where B is the batch size
    # query_feats: (B, N, D), where N is the number of queries
    # bbox: (B, N, 4), bounding box prediction made by previous decoder layer

    # Self-attention
    query_feats = self_attn(query_feats)  # (B, N, D)
    
    # Local Feature Sampling
    sparse_feats = RoIAlign(image_feats, bbox)  # (B, N, D, K, K)
    
    # For cross-attention, the batch size is treated as B*N for parallel decoding
    sparse_feats = sparse_feats.view(B*N, D, K*K).permute(0, 2, 1)  # (B*N, K*K, D)
    query_feats = query_feats.view(B*N, 1, D)
    query_feats = cross_attn(query_feats, sparse_feats).view(B, N, D)
    
    # Predictions in the current decoder layer
    class_probs = classifier(query_feats)
    bbox = regressor(query_feats)  # for feature sampling in the next decoder layer
    
    return query_feats, class_probs, bbox
    
\end{lstlisting}
\caption{Pseudo-code for the forward function of a single decoder layer with local feature sampling. We use a single-scale image feature for illustration. 
}
\label{fig:code}
\end{figure}
With the proposed local feature sampling in the decoder layer, each object query attend to different set of keys. To facilitate parallel decoding in the cross-attention layer, we treat different queries as individual samples in a batch.
Specifically, suppose a batch of input queries to the decoder has a shape of $(B, N, D)$, where $B$ is the batch size. In cross-attention, the queries are viewed as $(B\times N, 1, D)$ for parallel decoding. The pseudo-code in Fig.~\ref{fig:code} illustrate the forward function of a single decoder layer with local feature sampling.


\section{Discussions on Local Feature Sampling}
Our model transformation shows that sparse feature sampling from local areas is critical to data efficiency. 
Though PnP-DETR~\cite{pnp-detr} also attempts to sample sparse features, the features are sampled from the entire image, instead of a local area in the image. As a result, the sampled features may contain multiple instances and the model still has to learn to focus on specific objects from more training data. By contrast, Sparse RCNN, DeformDETR, and our method sample sparse features from local object regions, thus alleviating the data-hungry issue.

\section{DeformDETR with Label Augmentation}
Since DeformDETR already performs sample feature sampling from local areas and multi-scale feature fusion with the advanced Deformable Attention~\cite{zhu2020deformable}, we do not modify its model structure. Instead, we simply combine our label augmentation method with DeformDETR, which is denoted as LA-DeformDETR.

\begin{table}[ht]
\renewcommand{\arraystretch}{1.0}
\renewcommand{\tabcolsep}{2.pt}
\centering
\caption{DeformDETR trained with the proposed label augmentation (LA) on Cityscapes. ``SF", ``MS", and ``LA" represent sparse feature sampling, multi-scale feature fusion, and label augmentation, respectively.}
\label{table: app_deform_detr_city}
\resizebox{\textwidth}{!}{
    \begin{tabular}{l|ccc|cccccc|ccc}
    \Xhline{2\arrayrulewidth}  
    \hline
    \rowcolor{mygray}
    Method & SF & MS & LR & AP & AP$_\text{50}$ & AP$_\text{75}$ & AP$_\text{S}$ & AP$_\text{M}$ & AP$_\text{L}$ & \tabincell{c}{Params} & \tabincell{c}{FLOPs} & FPS \\
    \hline
    DeformDETR~\cite{zhu2020deformable} & $\checkmark$ & $\checkmark$ &  & 27.3 & 49.2 & 26.3 & 8.7 & 28.2 & 45.7 & 40M & 174G & 28 \\
    \hline
    LA-DeformDETR & $\checkmark$ & $\checkmark$ & $\checkmark$ & 28.6 & 52.2 & 27.4 & 8.9 & 28.9 & 47.9 & 40M & 174G & 28 \\
    \Xhline{2\arrayrulewidth}
    \end{tabular}
}
\end{table}
\noindent\textbf{Experiments on Cityscapes.} As can be seen from Table~\ref{table: app_deform_detr_city}, the DeformDETR is a data-efficient model that achieves an even better performance than Sparse RCNN on Cityscapes. However, our label augmentation can further improve its data efficiency and achieve a 1.3 AP gain to the strong baseline.

\begin{figure*}[t]
  \centerline{\includegraphics[width=7.5cm]{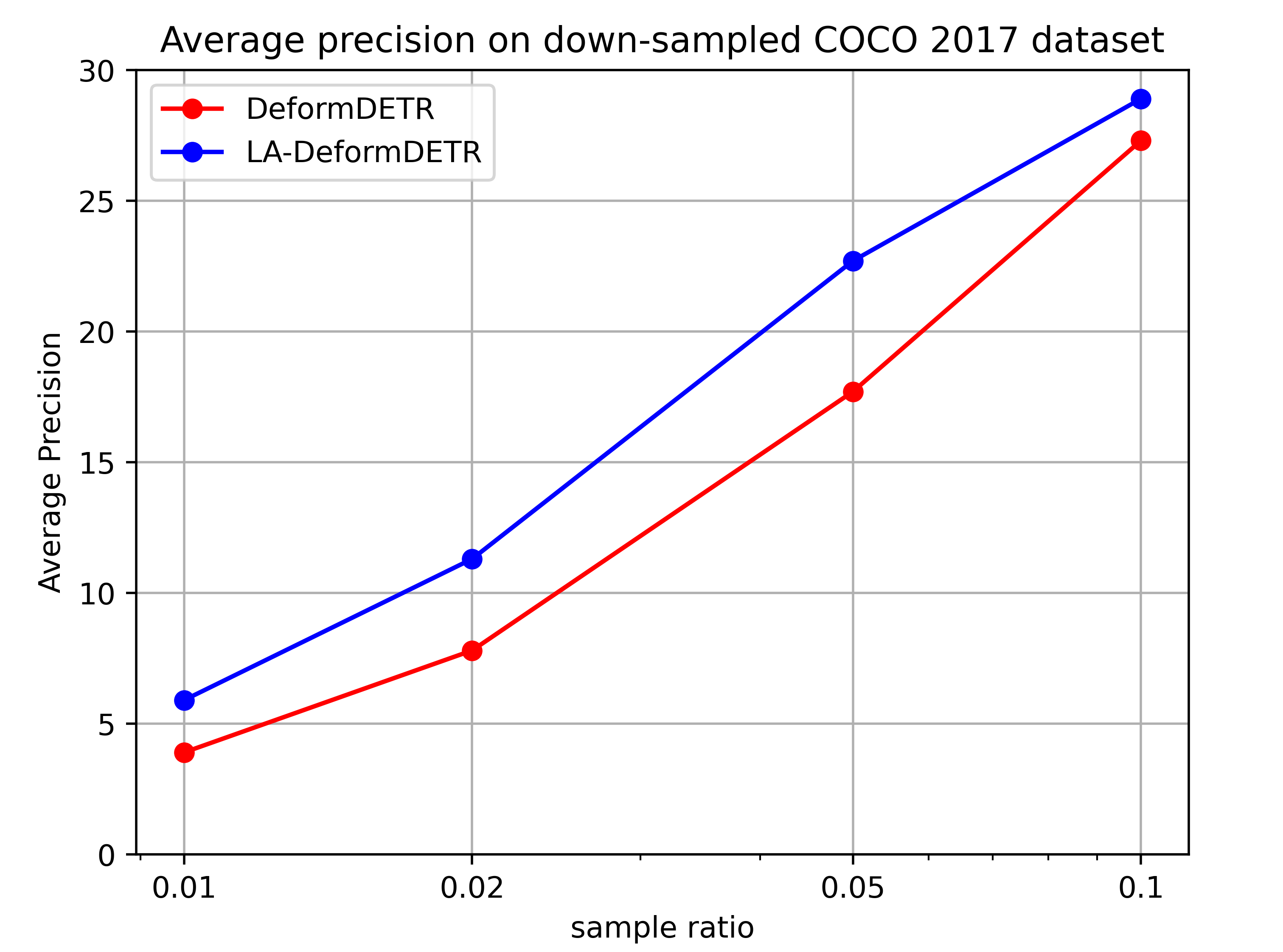}}
  \caption{Performance comparison of DeformDETR variants on sub-sampled COCO 2017 dataset. Note the sample ratio is shown on a logarithmic scale. As can be seen, the proposed label augmentation consistently improves the model performance under varying data sampling ratios.}
  \label{fig: deform_cocodown}
\end{figure*}
\noindent\textbf{Experiments on the sub-sampled COCO 2017.} We also evaluate the performance of our LA-DeformDETR on the sub-sampled COCO 2017 dataset. As shown in Fig.~\ref{fig: deform_cocodown}, our method consistently outperforms the DeformDETR baseline under varying sampling rates.

\section{Ablations based on DELA-DETR}
To gain a more comprehensive understanding of each component in our method, we also conduct ablation studies based on DELA-DETR, following the ablation studies in Section 5.2 of our main text.

\begin{table}[ht]
\renewcommand{\arraystretch}{1.0}
\renewcommand{\tabcolsep}{2.pt}
\centering
\caption{Ablations on each component in DELA-DETR. ``SF", ``MS", and ``LA" represent sparse feature sampling, multi-scale feature fusion, and label augmentation, respectively. $\dagger$ indicates the query number is increased from 100 to 300.}
\label{table: app_ablate_detr_main}
\resizebox{\textwidth}{!}{
    \begin{tabular}{l|c|ccc|cccccc|ccc}
    \Xhline{2\arrayrulewidth}  
    \hline
    \rowcolor{mygray}
    Method & Epochs & SF & MS & LA & AP & AP$_\text{50}$ & AP$_\text{75}$ & AP$_\text{S}$ & AP$_\text{M}$ & AP$_\text{L}$ & \tabincell{c}{Params} & \tabincell{c}{FLOPs} & FPS \\
    \hline
    DETR~\cite{detr} & 300 &  &  &  & 11.5 & 26.7 & 8.6 & 2.5 & 9.5 & 25.1 & 41M & 86G & 44 \\
    \hline
    LA-DETR & 300 &  &  & $\checkmark$ & 16.8 & 36.9 & 13.3 & 3.0 & 13.4 & 35.5 & 41M & 86G & 44 \\
    & 50 & $\checkmark$ &  &  & 16.3 & 34.9 & 12.9 & 2.3 & 12.0 & 35.9 & 42M & 85G & 36 \\
    \hline 
    DE-DETR & 50 & $\checkmark$ & $\checkmark$ &  & 21.7 & 41.7 & 19.2 & 4.9 & 20.0 & 39.9 & 42M & 88G & 34 \\
    DELA-DETR & 50 & $\checkmark$ & $\checkmark$ & $\checkmark$ & 20.6 & 40.1 & 18.4 & 4.6 & 18.9 & 37.5 & 42M & 91G & 29 \\
    DE-DETR$^\dagger$ & 50 & $\checkmark$ & $\checkmark$ &  & 22.4 & 41.3 & 20.9 & 6.0 & 21.3 & 39.7 & 42M & 91G & 29 \\
    DELA-DETR$^\dagger$ & 50 & $\checkmark$ & $\checkmark$ & $\checkmark$ & 24.5 & 46.2 & 22.5 & 6.1 & 23.3 & 43.9 & 42M & 91G & 29 \\
    \Xhline{2\arrayrulewidth}
    \end{tabular}
}
\end{table}

\noindent\textbf{Effectiveness of each module.} As shown in Table~\ref{table: app_ablate_detr_main}, both sparse feature sampling from local areas and multi-scale feature significantly improve the data efficiency of DETR, bringing a 4.8 and 5.4 gain on AP even under a much shorter training schedule. Besides, we also apply label augmentation to DE-DETTR. However, directly applying label augmentation makes the performance worse. We conjecture that with only 100 queries, the positive sample ratio becomes too high for training, particularly for the Cityscapes dataset, where many images contain dense object scenes. To solve this problem, we improve the number of queries from 100 to 300, as did in CondDETR and DeformDETR. As can be seen, DELA-DETR$^\dagger$ outperforms DE-DETR by 2.8 AP. Moreover, from the comparison between DE-DETR$^\dagger$ and DELA-DETR$^\dagger$, it can be seen that the performance gain mainly comes from our label augmentation, instead of more queries.

\begin{table}[ht]
\renewcommand{\arraystretch}{1.0}
\renewcommand{\tabcolsep}{2.pt}
\centering
\caption{Ablations on multi-scale feature levels and feature resolutions for RoIAlign, experimented based on DE-DETR. Note label augmentation is not utilized for clarity.}
\label{table: app_ablate_detr_ms_res}
    \begin{tabular}{cc|cccccc|ccc}
    \Xhline{2\arrayrulewidth}  
    \hline
    \rowcolor{mygray}
    MS Lvls & RoI Res. & AP & AP$_\text{50}$ & AP$_\text{75}$ & AP$_\text{S}$ & AP$_\text{M}$ & AP$_\text{L}$ & \tabincell{c}{Params} & \tabincell{c}{FLOPs} & FPS \\
    
    \hline
    1 & 1 & 12.5 & 30.8 & 8.1 & 1.8 & 8.9 & 26.9 & 42M & 85G & 36 \\
    1 & 4 & 16.3 & 34.9 & 12.9 & 2.3 & 12.0 & 35.9 & 42M & 85G & 36 \\
    1 & 7 & 16.5 & 35.6 & 13.0 & 2.3 & 12.5 & 36.5 & 42M & 86G & 36 \\
    \hline
    3 & 4 & 21.7 & 41.7 & 19.2 & 4.9 & 20.0 & 39.9 & 42M & 88G & 34 \\
    4 & 4 & 21.1 & 40.8 & 18.9 & 4.3 & 18.6 & 39.5 & 47M & 89G & 33 \\
    \Xhline{2\arrayrulewidth}
    \end{tabular} 
\end{table}

\noindent\textbf{Resolution for RoIAlign and number of multi-scale features.} As can be seen from Table~\ref{table: app_ablate_detr_ms_res}, RoIAlign with a feature resolution of 4 is both efficient and effective. And the three feature levels for multi-scale fusion achieve the optimal performance. Thus, the hyper-parameter settings of DE-DETR are exacted the same as that of DE-CondDETR.

\begin{table}[ht]
\renewcommand{\arraystretch}{1.0}
\renewcommand{\tabcolsep}{2.pt}
\centering
\caption{Ablation on the proposed label augmentation, experimented based on DE-DETR. Params, FLOPs, and FPS are omitted since they are consistent for all label augmentation settings.}
\label{table: app_ablate_detr_label_aug}
    \begin{tabular}{cc|cccccc}
    \Xhline{2\arrayrulewidth}  
    \hline
    \rowcolor{mygray}
    Fix Time & Fix Ratio & AP & AP$_\text{50}$ & AP$_\text{75}$ & AP$_\text{S}$ & AP$_\text{M}$ & AP$_\text{L}$ \\
    \hline
    -- & -- & 21.7 & 41.7 & 19.2 & 4.9 & 20.0 & 39.9 \\
    \hline
    2 & -- & 24.5 & 46.2 & 22.5 & 6.1 & 23.3 & 43.9 \\
    3 & -- & 23.7 & 43.8 & 21.8 & 5.7 & 21.7 & 43.5 \\
    4 & -- & 22.8 & 42.7 & 21.1 & 5.7 & 21.3 & 41.5 \\
    5 & -- & 22.6 & 43.1 & 20.1 & 5.5 & 20.7 & 40.9 \\
    \hline
    -- & 0.1 & 23.8 & 44.8 & 21.1 & 5.6 & 22.4 & 42.0 \\
    -- & 0.2 & 23.4 & 43.7 & 21.2 & 6.0 & 21.7 & 41.8 \\
    -- & 0.25 & 23.1 & 43.4 & 21.0 & 5.4 & 22.1 & 41.1 \\
    -- & 0.3 & 23.1 & 43.3 & 21.1 & 5.7 & 21.8 & 41.0 \\
    -- & 0.4 & 21.9 & 40.4 & 20.5 & 5.1 & 21.0 & 39.8 \\
    \Xhline{2\arrayrulewidth}
    \end{tabular}
\end{table}

\noindent\textbf{Ablations on label augmentation.} We also ablate the proposed label augmentation method with the fixed repeat time strategy and the fixed positive sample ratio strategy. As can be seen from Table~\ref{table: app_ablate_detr_label_aug}, a fixed repeat time of 2 consistently achieve the best performance.

\section{Ablations on NMS}
Since the proposed label augmentation performs one-to-many matching between ground truths and predictions, a duplicate remove process is required. Although more advanced duplicate removal methods, like Soft-NMS~\cite{soft-nms}, can be used, we simply adopt the vanilla NMS.

\begin{table}[ht]
\renewcommand{\arraystretch}{1.0}
\renewcommand{\tabcolsep}{4.pt}
\centering
\caption{Ablations on NMS, experiments based on DELA-CondDETR. $N_t$ indicates the IoU threshold for NMS.}
\label{table: app_ablate_nms}
    \begin{tabular}{l|c|cccccc}
    \Xhline{2\arrayrulewidth}  
    \hline
    \rowcolor{mygray}
    Method & $N_t$ & AP & AP$_\text{50}$ & AP$_\text{75}$ & AP$_\text{S}$ & AP$_\text{M}$ & AP$_\text{L}$ \\
    \hline
    \multirow{2}{*}{DE-CondDETR} & -- & 26.8 & 47.8 & 25.4 & 6.8 & 25.6 & 46.6 \\
    & 0.7 & 27.1 & 49.2 & 25.3 & 6.8 & 25.9 & 47.5 \\
    \hline 
    \multirow{7}{*}{DELA-CondDETR} & 0.3 & 28.6 & 51.9 & 26.6 & 7.1 & 27.3 & 49.3 \\
    & 0.4 & 29.0 & 52.7 & 26.9 & 7.2 & 27.6 & 49.7 \\
    & 0.5 & 29.3 & 53.2 & 27.1 & 7.3 & 28.0 & 50.0 \\
    & 0.6 & 29.5 & 53.2 & 27.4 & 7.5 & 28.2 & 50.1 \\
    & 0.7 & 29.5 & 52.8 & 27.6 & 7.5 & 28.2 & 50.1 \\
    & 0.8 & 29.3 & 51.8 & 27.8 & 7.4 & 28.0 & 49.8 \\
    & 0.9 & 28.5 & 49.6 & 27.5 & 7.1 & 26.9 & 48.7 \\
    
    \Xhline{2\arrayrulewidth}
    \end{tabular}
\end{table}

\noindent\textbf{NMS with different IoU thresholds.} We first ablate NMS with different IoU thresholds. As can be seen from Table~\ref{table: app_ablate_nms}, DELA-CondDETR with different IoU thresholds for NMS consistently outperforms the DE-CondDETR. Moreover, NMS with a wide range of IoU thresholds from 0.4 to 0.8 can achieve good performance. We conjecture that it is easier to remove the duplicates when each positive sample is repeated only two times. 

\noindent\textbf{DE-CondDETR with NMS.} As can be seen from Table~\ref{table: app_ablate_nms}, applying NMS on DE-CondDETR does not make much difference on model performance, since it follows a one-to-one matching scheme during training. This validates the performance gain of DELA-CondDETR comes from a richer supervision signal, instead of the duplicate removal process.

\section{Ablations on the Box Refinement}
\begin{table}[ht]
\renewcommand{\arraystretch}{1.0}
\renewcommand{\tabcolsep}{2.pt}
\centering
\caption{Ablations on the single-scale sparse feature sampling, experimented based on CondDETR. RoI and Refine indicate RoIALign and cascaded bounding box refinement, respectively. All models are trained for 50 epochs.}
\label{table: app_ablate_refine_cond}
    \begin{tabular}{l|cc|cccccc|ccc}
    \Xhline{2\arrayrulewidth}  
    \hline
    \rowcolor{mygray}
    Method & Refine & RoI & AP & AP$_\text{50}$ & AP$_\text{75}$ & AP$_\text{S}$ & AP$_\text{M}$ & AP$_\text{L}$ & \tabincell{c}{Params} & \tabincell{c}{FLOPs} & FPS \\
    \hline
    CondDETR~\cite{meng2021conditional} &  &  & 12.1 & 28.0 & 9.1 & 2.2 & 9.8 & 27.0 & 43M & 90G & 39 \\
    \hline
    & $\checkmark$ &  & 12.1 & 30.2 & 7.8 & 2.1 & 10.4 & 25.7 & 44M & 90G & 39 \\
    &  & $\checkmark$ & 16.3 & 32.5 & 14.3 & 3.3 & 15.0 & 33.6 & 43M & 95G & 32  \\
    \hline
    & $\checkmark$ & $\checkmark$ & 20.4 & 40.7 & 17.7 & 2.9 & 16.9 & 42.0 & 44M & 95G & 32 \\
    \Xhline{2\arrayrulewidth}
    \end{tabular} 
\end{table}
For sparse feature sampling from local areas, both RoIAlign and cascaded bounding box refinement are included. In this section, we ablate the effectiveness of these two components. As shown in Table~\ref{table: app_ablate_refine_cond}, RoIAlign alone can bring a 4.2 gain on AP. By contrast, applying the bound box refinement alone does not improve the model performance. However, when combined with RoIAlign, the bound box refinement can improve the model performance by 4.1 AP.
Since the regression targets for each decoder layer are determined by the bounding box predictions made by the previous layer, we conjecture that it is important that the current decoder layer is aware of the updated reference boxes, in order to make the box refinement effective. 
With RoIAlign, the current decoder layer can infer the updated reference boxes from the sampled features, which have been added with the sine positional embedding. By contrast, when applying the bounding boxes refinement alone, the updated reference boxes can hardly be inferred from the global feature map of the entire image.

\section{Experiments on Pascal VOC}
\begin{table}[ht]
\renewcommand{\arraystretch}{1.0}
\renewcommand{\tabcolsep}{2.pt}
\centering
\caption{Performance of our data-efficient detection transformers on Pascal VOC. All models are trained on trainval07+12 for 50 epochs, and evaluated on test2007.}
\label{table: voc}
    \begin{tabular}{l|c|cccccc|ccc}
    \Xhline{2\arrayrulewidth}  
    \hline
    \rowcolor{mygray}
    Method & Epochs & AP & AP$_\text{50}$ & AP$_\text{75}$ & AP$_\text{S}$ & AP$_\text{M}$ & AP$_\text{L}$ & \tabincell{c}{Params} & \tabincell{c}{FLOPs} & FPS \\
    \hline
    DETR~\cite{detr} & 50 & 38.3 & 62.1 & 40.3& 2.1 & 12.9 & 48.3 & 41M & 86G & 43 \\
    DE-DETR & 50 & 54.8 & 78.7 & 60.7 & 20.2 & 39.2 & 61.6 & 43M & 88G & 33 \\
    DELA-DETR$^\dagger$ & 50 & 57.0 & 82.0 & 63.0 & 24.5 & 42.1 & 63.8 & 43M & 91G & 29 \\
    \hline
    CondDETR~\cite{meng2021conditional} & 50 & 55.6 & 82.0 & 60.9 & 15.1 & 34.7 & 63.9 & 43M & 90G & 39 \\
    DE-CondDETR & 50 & 56.4 & 80.2 & 63.2 & 22.2 & 40.7 & 62.8 & 44M & 107G & 28 \\
    DELA-CondDETR & 50 & 59.5 & 84.4 & 66.3 & 29.7 & 43.6 & 65.6 & 44M & 107G & 28 \\
    \Xhline{2\arrayrulewidth}
    \end{tabular}
\end{table}
In this section, we also conducted experiments on based on the Pascal VOC dataset~\cite{voc}. Follow the common practice, we train models on the trainval07+12 split with contains about 16.5k images and evaluated them on
the test2007 split. All models are trained with a batch size of 32. The results are shown in Table~\ref{table: voc}, as can be seen, both our DE- and DELA- model variants consistently outperform the corresponding baselines.

\section{Visualizations}

\begin{figure*}
  \centerline{\includegraphics[width=\textwidth]{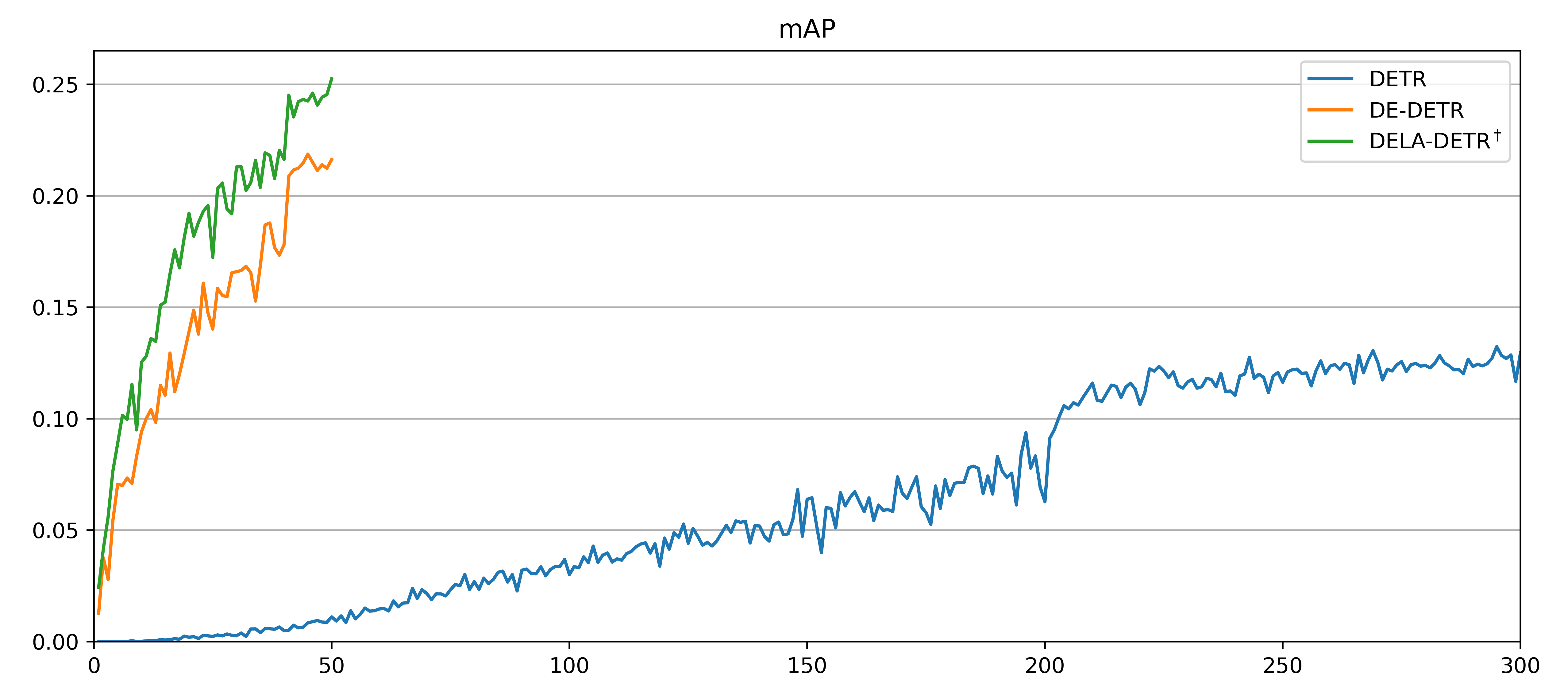}}
  \caption{Convergence curves of DETR variants on Cityscapes.}
  \label{fig:convergence_detr}
\end{figure*}

\begin{figure*}
  \centerline{\includegraphics[width=7.5cm]{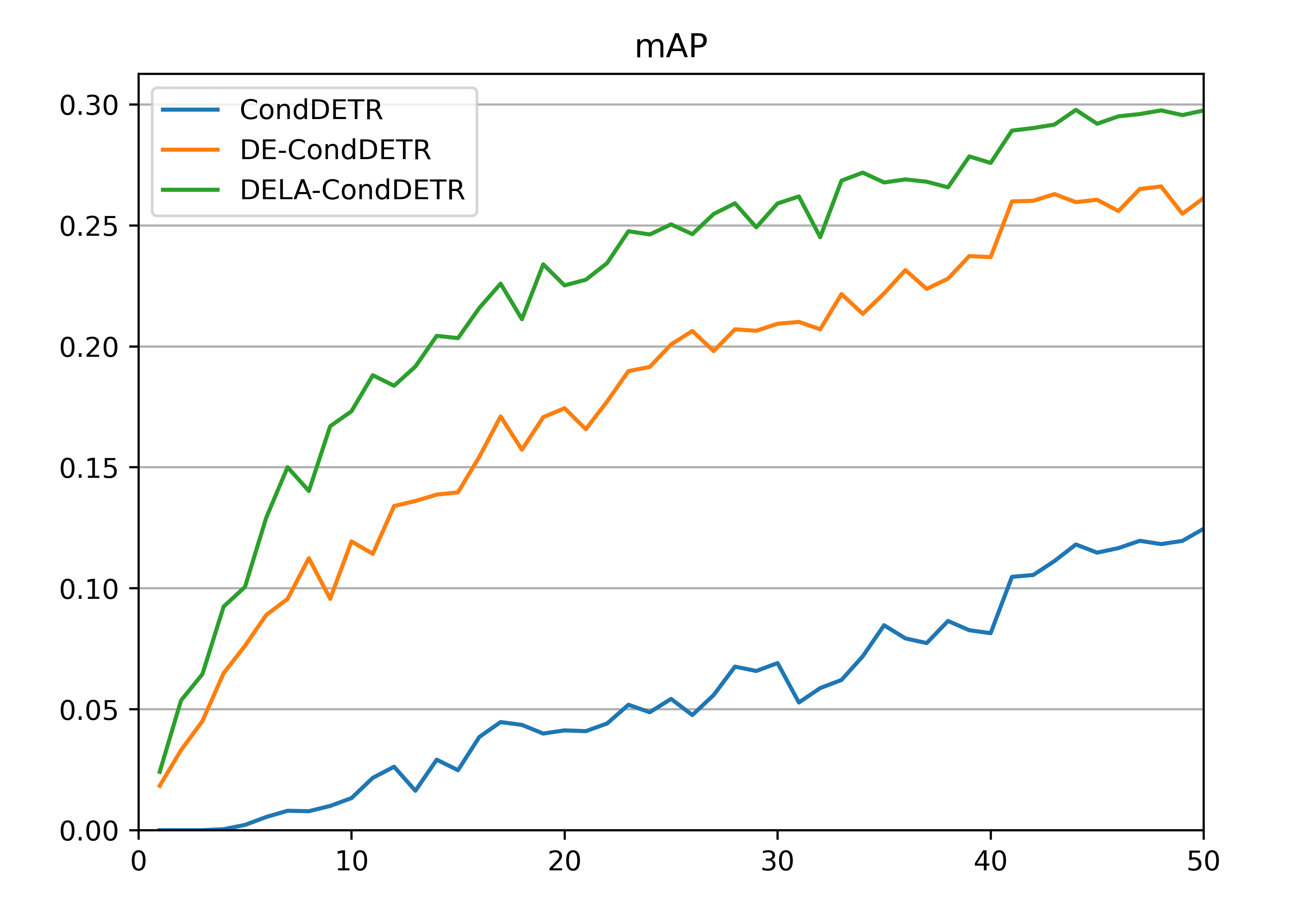}}
  \caption{Convergence curves of CondDETR variants on Cityscapes.}
  \label{fig:convergence_cond_detr}
\end{figure*}

\noindent\textbf{Visualization of training curves.}
The comparison of DETR variants is shown in Fig.~\ref{fig:convergence_detr}. As can be seen, both DE-DETR and DELA-DETR$^\dagger$ can achieve better performance and faster convergence compared with the DETR baseline. Similarly, the comparison of ConDETR variants in Fig.~\ref{fig:convergence_cond_detr} also validates the effectiveness of our method.

\begin{figure*}
  \centerline{\includegraphics[width=\textwidth]{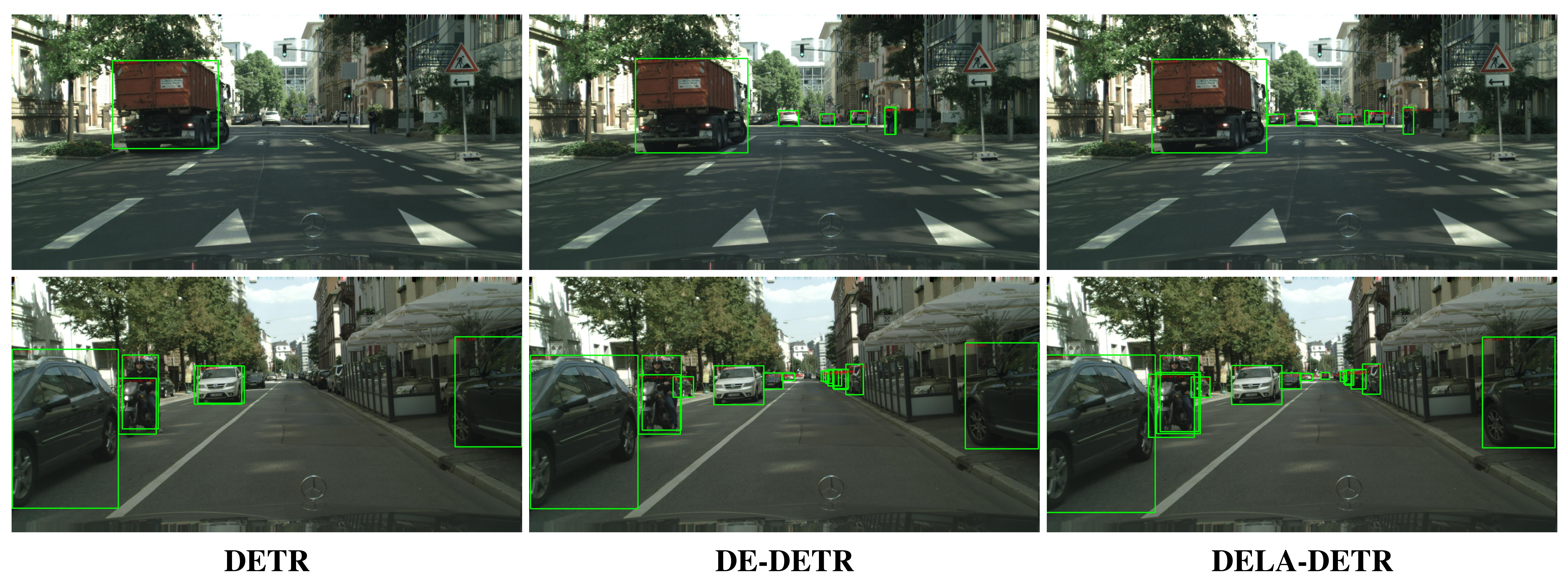}}
  \caption{Demo detection results of DETR variants on Cityscapes.}
  \label{fig:demo_detr}
\end{figure*}

\begin{figure*}
  \centerline{\includegraphics[width=\textwidth]{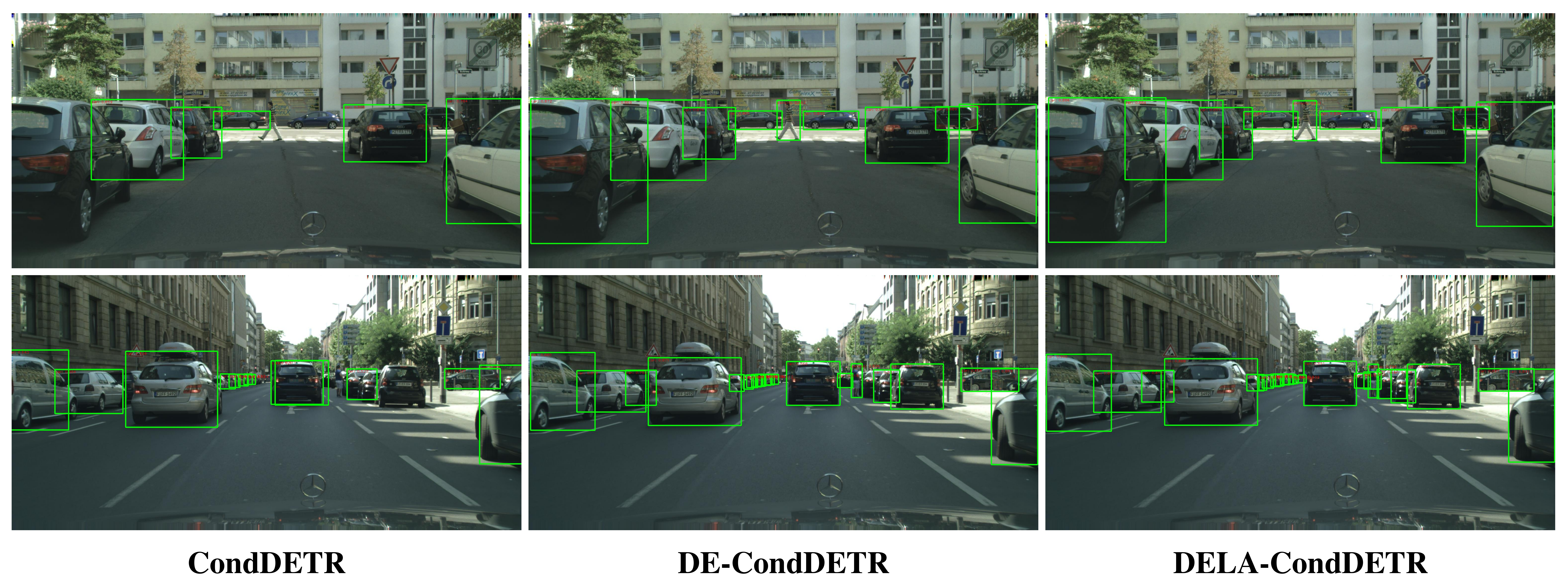}}
  \caption{Demo detection results of CondDETR variants on Cityscapes.}
  \label{fig:demo_cond_detr}
\end{figure*}

\noindent\textbf{Visualization demo results.}
We also provide demo detection results of different methods on the Cityscapes dataset, as shown in Fig.~\ref{fig:demo_detr} and Fig.~\ref{fig:demo_cond_detr}. It can be seen that our method can detect the distant instances overlooked by the baseline methods. Moreover, they often avoid the false positive prediction made by the baseline methods, as shown in the second rows in both Fig.~\ref{fig:demo_detr} and Fig.~\ref{fig:demo_cond_detr}.


\section{Discussions on limitations}
Though effective, our method may have certain potential disadvantages:
(1) A small and fixed resolution for feature sampling may be detrimental to the detection of large objects, as can be seen from the comparison between CondDETR and DELA-CondDETR in Table 7. This can be alleviated by more effective techniques that adaptively sample more feature points for large objects. (2) The sampling process may slow down the inference speed, which can be alleviated by a joint CUDA implementation of both the feature sampling and the following cross-attention processes.

\section{Future Work}
\noindent\textbf{Data efficiency of transformers on different vision tasks.}
With limited inductive bias, the vision transformers are often data-hungry~\cite{dosovitskiy2020image}. Although the data-hungry issue of vision transformer for image classification has been studied~\cite{deit}, it has not been explored for other vision tasks. In this paper, we take the first step to delve into the data-hungry issue of detection transformers. As transformers become increasingly popular for vision tasks, like semantic segmentation~\cite{zheng2021rethinking,xie2021segformer}, 3D object detection~\cite{misra2021end}, and video instance segmentation~\cite{wang2021end}, we hope our work will inspire the community to explore the data efficiency of transformers for different tasks.

\noindent\textbf{Removing NMS.}
The one-to-many matching between ground-truths and predictions in our label augmentation can provide richer training supervision to alleviate the data-hungry issue. However, it also brings the need for the duplicate removal process. Though the NMS has been proved effective, we hope to remove this post-processing step to maintain the end-to-end property of detection transformers. To achieve this, we provide three possible solutions as follows. Firstly, a lightweight duplicate removal network can be trained along with the model, as did in Relation Network~\cite{hu2018relation}. Secondly, an additional rank loss~\cite{liu2021rankdetnet} can be applied to regularize the score of the predictions, so that the predictions matched to the original ground-truths can rank higher. In this way, the duplicate removal process is no longer needed. Thirdly, we can apply the label augmentation only on the shallower decoder layers while keeping the supervision on the deep decoder layers unchanged. In this way, the queries at the shallower decoder layers can receive a rich supervision signal, while the deeper decoder layers only select the most promising query for each target instance.

\noindent\textbf{Further improving the data efficiency of existing detection transformers.}
Although outperforming all existing detection transformers, there is still a gap between the data efficiency of our method and the seminal Faster-RCNN-FPN. To bridge this gap, a possible solution is to gradually transform a Faster-RCNN-FPN to a Sparse RCNN, to find the key reasons for Faster-RCNN-FPN's data efficiency. Afterward, we can adjust the detection transformer structures accordingly to improve their data efficiency.

\end{document}